\documentclass{article}
\usepackage{loopmoevr,times}
\usepackage{amsmath,amssymb}
\usepackage{graphicx,booktabs,array,adjustbox,multirow}
\usepackage[table]{xcolor}
\usepackage{fontspec}
\usepackage[unicode,colorlinks=true,citecolor=blue,linkcolor=black,urlcolor=black]{hyperref}
\DeclareRobustCommand{\visualref}[1]{\hyperref[#1]{\textcolor{red}{\ref*{#1}}}}
\usepackage{url,newunicodechar}
\usepackage{placeins,tabularx,needspace}
\usepackage{capt-of,pifont}
\newunicodechar{Δ}{\ensuremath{\Delta}}
\newunicodechar{∈}{\ensuremath{\in}}
\newunicodechar{→}{\ensuremath{\rightarrow}}
\newunicodechar{↑}{\ensuremath{\uparrow}}
\newunicodechar{↓}{\ensuremath{\downarrow}}
\newunicodechar{×}{\ensuremath{\times}}
\providecommand{\tightlist}{\setlength{\itemsep}{2pt}\setlength{\parskip}{0pt}}
\newcommand{\ablationtile}[4]{%
\begin{minipage}[t]{#1}\centering
{\scriptsize (#2)}\par\vspace{1pt}
\includegraphics[width=\linewidth]{#3}\par\vspace{1pt}
\parbox[c][17pt][c]{\linewidth}{\centering\fontsize{6.8}{7.6}\selectfont #4}%
\end{minipage}}
\newlength{\introfigheight}
\title{\raggedright LoopMoEVR: Loop-Based Degradation-Aware Mixture-of-Experts for Unified UHD Video Restoration}
\author{%
\begin{minipage}[t]{\dimexpr\textwidth-2\tabcolsep\relax}\centering
\normalsize\bfseries
Yucheng Xin\textsuperscript{1,*},\; Runci Bai\textsuperscript{2},\; Yongcong Wang\textsuperscript{3},\; Guangwei Gao\textsuperscript{4},\\
Jiao Liu\textsuperscript{5},\; Dianjie Lu\textsuperscript{1},\; Linwei Fan\textsuperscript{6},\; Zhuoran Zheng\textsuperscript{7,\ensuremath{\dagger}}\\[3pt]
\fontsize{8}{9.5}\selectfont\normalfont
\textsuperscript{1}Shandong Normal University\quad
\textsuperscript{2}China Academy of Information and Communications Technology\\
\textsuperscript{3}Central South University\quad
\textsuperscript{4}Nanjing University of Science and Technology\\
\textsuperscript{5}Hebei University of Technology\quad
\textsuperscript{6}Shandong University of Finance and Economics\\
\textsuperscript{7}National University of Defense Technology
\end{minipage}}
\begin{document}
\maketitle
\fancyhead{}
\renewcommand{\headrulewidth}{0pt}
\begingroup
\renewcommand{\thefootnote}{\fnsymbol{footnote}}
\footnotetext[1]{\href{mailto:202411000525@stu.sdnu.edu.cn}{202411000525@stu.sdnu.edu.cn}}
\footnotetext[2]{Corresponding author: \href{mailto:zhengzr@njust.edu.cn}{zhengzr@njust.edu.cn}}
\endgroup

\begin{abstract}
Recently, unified high-definition image restoration has attracted considerable attention; however, existing models tend to excessively increase their depth in pursuit of improved generalization, which often yields only limited gains. Meanwhile, loop-based learning paradigms have drawn widespread attention due to their low parameter counts and strong regression capability, as exemplified by GPT-6 and looped Transformers. In this paper, we introduce the loop learning paradigm to address restoration tasks that require cross-domain learning. Specifically, we propose LoopMoEVR, a loop-based mixture-of-experts model capable of handling degraded ultra-high-definition (UHD) inputs. First, a degradation-conditioned low-rank loop embedding is designed to construct input-dependent stage conditions. Second, a spatio-temporal iterative adaptive normalization module, termed IterAda3DN, is developed to fuse local features with global loop context, thereby performing position-wise affine modulation. Finally, the expert branches further integrate the attention-updated local and global video states with the loop conditions to generate dedicated modulation parameters, while an input-conditioned depth predictor adaptively configures the number of loop iterations. With only approximately 0.884M trainable parameters, the proposed model uniformly handles UHD video dehazing, deraining, denoising, and low-light enhancement tasks, achieving state-of-the-art restoration performance on both public benchmarks and real-world scenarios.
\end{abstract}

\section{Introduction}\label{introduction}

Unified video restoration aims to address multiple degradations with a single model while exploiting cross-frame information to recover details and maintain temporal consistency. Extending unified restoration to ultra-high-definition (UHD) videos introduces further challenges. A key challenge is to increase restoration depth while retaining the ability to adapt to different degradations and evolving feature states.

For unified UHD video restoration, MoCRA~\citep{wang2026mocramixturecompositionalrank1} introduces UHV-4K-AIO, built from 100 clean 4K clips. It allocates conditioning capacity between a coarse branch and a full-resolution refiner according to degradation frequency, enabling native 4K restoration without optical flow using approximately 3.6M parameters. However, its frame-wise conditions do not iteratively refine intermediate representations, its fixed inference depth cannot adapt to degradation severity, and it does not explicitly model global video states or cross-iteration context. Unified UHD restoration therefore still calls for progressive refinement and adaptive depth within a limited parameter budget.

\begin{figure}[!t]
\centering
\setlength{\introfigheight}{.40\linewidth}
\begin{minipage}[t]{0.65\linewidth}
\vspace{0pt}
\centering
\includegraphics[width=\linewidth,height=\introfigheight,keepaspectratio]{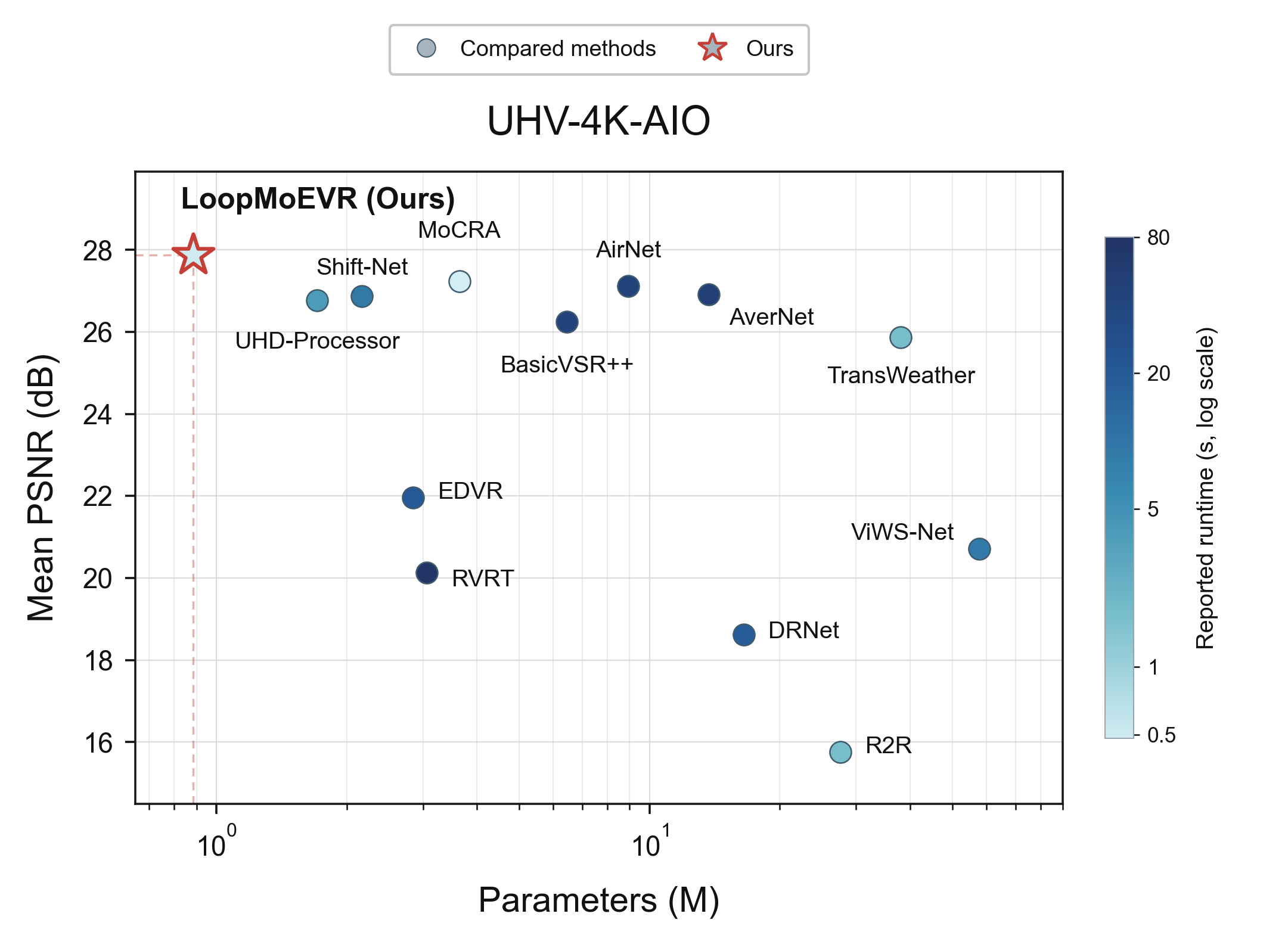}
\captionof{figure}{Mean PSNR, parameter count, and reported inference time on UHV-4K-AIO.}
\label{fig:intro-tradeoff}
\end{minipage}\hfill
\begin{minipage}[t]{0.32\linewidth}
\vspace{0pt}
\centering
\includegraphics[width=\linewidth,height=\introfigheight,keepaspectratio]{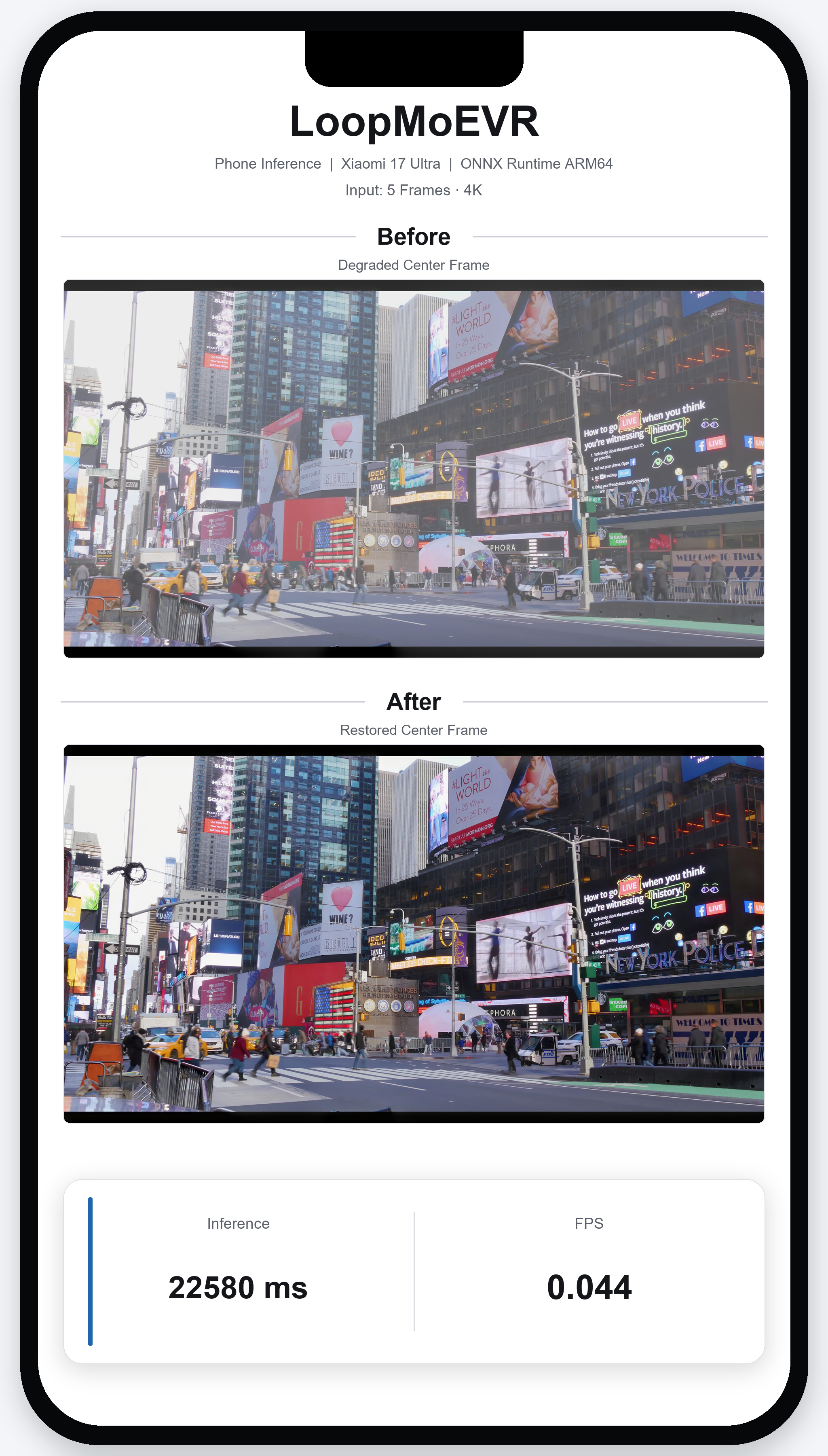}
\captionof{figure}{4K video restoration on a smartphone.}
\label{fig:intro-mobile}
\end{minipage}
\end{figure}

The loop learning paradigm offers a way to address these challenges by progressively refining restoration results over multiple iterations. The Universal Transformer~\citep{dehghani2018universal} repeatedly applies self-attention transformations along depth, supporting variable computation steps with a fixed parameter set. Subsequent studies examine the expressivity of looped computation and the role of iteration-dependent modulation \citep{saunshi2025reasoning,pmlr-v267-xu25x}. SMELT further investigates looped MoE models under matched resource budgets, demonstrating the effectiveness of looped computation for language modeling \citep{wang2026smelt}. LoopMoE combines looped computation with sparse experts and generates modulation parameters from iteration position and current local hidden states through IterAdaLN \citep{chen2026loopmoeunifyingiterativecomputation}. However, its base iteration embeddings are determined by the iteration index without explicit input-degradation conditions, and its modulation pathways do not directly integrate global video states or cross-iteration context. In unified UHD video restoration, the same iteration position may correspond to different restoration needs under different degradations, while spatio-temporal variation within a video requires local updates to incorporate clip-level state and loop context. We therefore construct a conditioning mechanism driven jointly by degradation, iteration stage, and video state, while aligning expert modulation with local features updated by attention, to support iterative restoration across degradation conditions.

To this end, we propose LoopMoEVR, a loop-based mixture-of-experts model for unified UHD video restoration that incorporates degradation conditions and video-state information into iterative restoration. The model comprises input encoding, looped latent-feature updates, and output reconstruction. First, degradation-conditioned low-rank loop embeddings combine input degradation representations, a severity-related scalar, and iteration position to generate input-dependent conditions for each restoration stage. Second, the spatio-temporal iterative adaptive normalization module, IterAda3DN, integrates local spatio-temporal features, global video states, and cross-iteration context to modulate features at each position. The expert branch further uses local and global video states updated by attention to condition expert transformations. In addition, a depth predictor selects the number of iterations according to the input, adapting the restoration process to different computational needs. By repeatedly applying spatio-temporal processing layers, LoopMoEVR performs multi-stage restoration with approximately 0.884M parameters. At the same network width and four-stage execution depth, it uses 71.6\% fewer parameters than a model with separate backbone parameters for each stage, with a PSNR difference of only 0.08 dB. This result shows that loop-based refinement achieves restoration quality close to that of stage-specific parameterization with a substantially more compact model, supporting our choice of a loop-based restoration architecture.
Figure~\visualref{fig:intro-tradeoff} compares the mean PSNR, parameter count, and reported inference time of LoopMoEVR and competing methods. Figure~\visualref{fig:intro-mobile} shows a smartphone-based 4K video restoration example.

\FloatBarrier

\Needspace{7\baselineskip}
Our main contributions are as follows:

\begin{enumerate}
\def\labelenumi{\arabic{enumi}.}
\tightlist
\item
  We propose LoopMoEVR for unified UHD video restoration, introducing the loop learning paradigm to progressively refine video features. By conditioning loop embeddings on input degradation representations, severity information, and iteration position, the model adapts feature updates to different degradation conditions and restoration stages.
\item
  We design IterAda3DN to integrate local spatio-temporal features with global loop context for position-wise affine modulation. Local and global video states updated by attention further provide dedicated modulation parameters for the expert branch, while an input-conditioned depth predictor assigns an appropriate loop depth to each input.
\item
  With approximately 0.884M trainable parameters, LoopMoEVR jointly addresses UHD video dehazing, deraining, denoising, and low-light enhancement. It achieves the highest mean PSNR on UHV-4K-AIO and performs well in evaluations of time-varying degradations and temporal consistency.
\end{enumerate}

\section{Related Work}\label{related-work}

\subsection{Unified Video Restoration}\label{unified-video-restoration}\label{degradation-aware-restoration}

Unified video restoration combines temporal information with a shared model across degradations. AverNet~\citep{zhao2024avernet}, UniWRV~\citep{Wan_2026}, and ORCANet~\citep{li2026allinonevideorestorationsmoothly} use alignment, weather priors, and clip- or frame-level prompts, respectively; MoCRA~\citep{wang2026mocramixturecompositionalrank1} targets native 4K restoration through frequency-aware, rank-one conditioning. In degradation-aware image restoration, AirNet~\citep{Li_2022_CVPR}, PromptIR~\citep{NEURIPS2023_e187897e}, and UtilityIR~\citep{chen2025always} adapt features through learned representations or prompts; AdaIR~\citep{cui2025adair} uses frequency cues, while MoCE-IR~\citep{zamfir2025complexity} and UniRestorer~\citep{lin2026unirestorer} study expert selection. UniLDiff~\citep{Cheng_2026_CVPR} incorporates degradation features into denoising, RONIN~\citep{janjua2026grounding} uses language-based degradation semantics, and video methods such as CDUN~\citep{cheng2023crossconsistentdeepunfoldingnetwork} and LaverNet~\citep{zhao2025lavernetlightweightallinonevideo} study adaptive updates and selective propagation. These approaches motivate input-aware restoration; our focus is degradation-conditioned, parameter-shared refinement of UHD video.

\subsection{Looped Deep Learning}\label{looped-deep-learning}

Looped models increase effective depth by reusing computation, as in the Universal Transformer~\citep{dehghani2018universal}, Huginn~\citep{geiping2025scalingtesttimecomputelatent}, ELT~\citep{goyal2026elt}, and Looped World Models~\citep{lu2026loopedworldmodels}. Studies of loop expressivity~\citep{saunshi2025reasoning} and timestep encoding~\citep{pmlr-v267-xu25x} examine repeated transformations, while LoopFormer~\citep{jeddi2026loopformer} conditions updates on iteration step size and Mixture-of-Recursions~\citep{bae2025mixtureofrecursions} routes token-level depth. LoopMoE~\citep{chen2026loopmoeunifyingiterativecomputation} combines sparse experts with iteration- and local-state modulation; its expert branch already uses the attention-updated state. SMELT~\citep{wang2026smelt} evaluates looped MoE under matched resource budgets. Building on this work, LoopMoEVR conditions loop embeddings on input degradation, incorporates global video state and cross-iteration context into modulation, and selects depth according to the input.

\section{Method}\label{method}

\subsection{Overall Framework}\label{overall-framework}

\begin{figure}[!t]
\centering
\includegraphics[width=.86\linewidth]{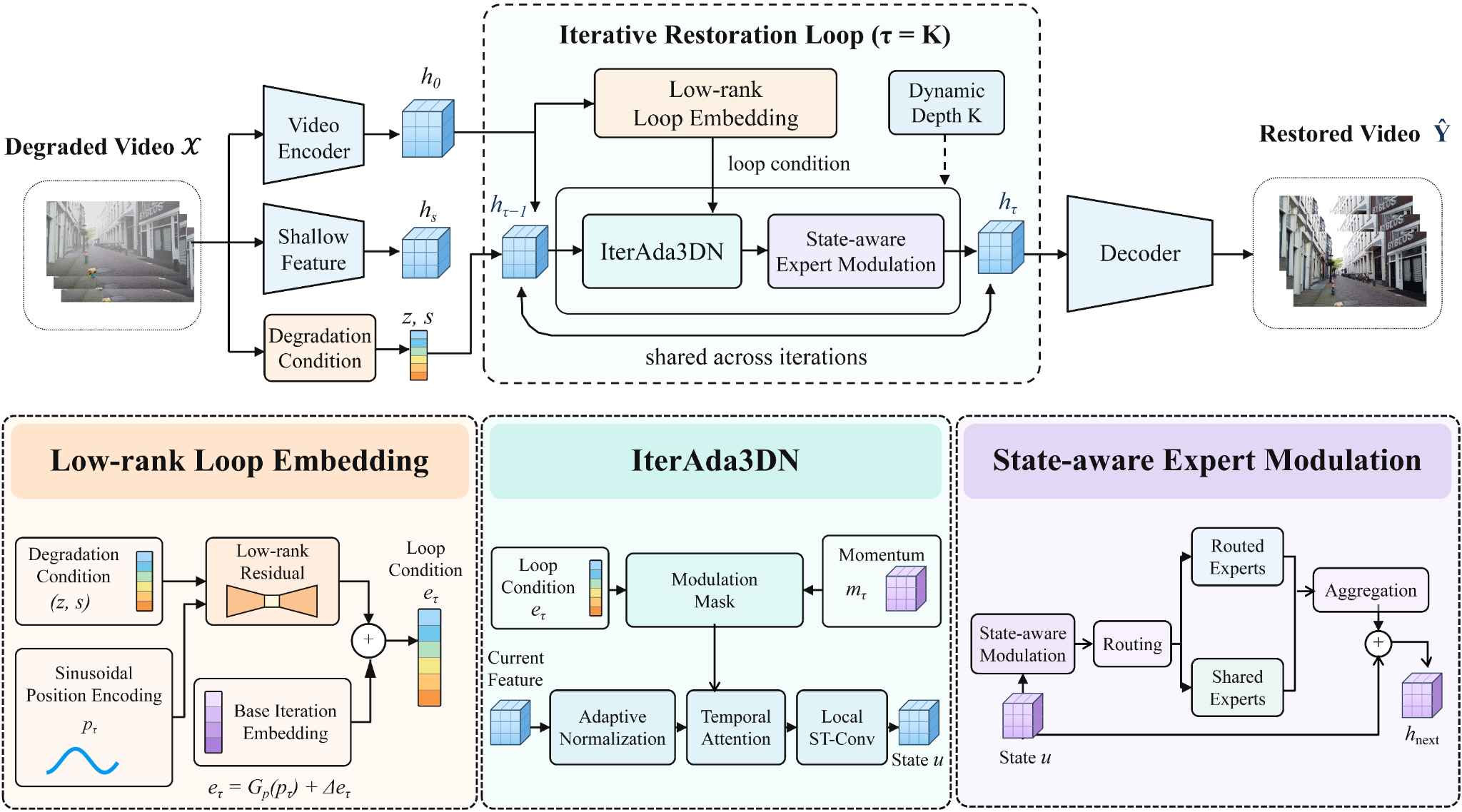}
\caption{\textbf{Iterative restoration mechanism of LoopMoEVR.}
Degradation-conditioned low-rank loop embedding guides IterAda3DN
and state-aware expert modulation to progressively refine video
features within a parameter-shared loop.}
\label{fig:overall_framework}
\end{figure}

Figure~\visualref{fig:overall_framework} illustrates the overall mechanism of LoopMoEVR.
Following the shared-computation setting of LoopMoE~\citep{chen2026loopmoeunifyingiterativecomputation}, given a degraded video clip \(\mathbf X\), LoopMoEVR extracts latent features and degradation conditions, then iteratively updates video features through degradation-conditioned loop embeddings, spatio-temporal iterative adaptive normalization, and state-aware expert modulation. An input-conditioned depth predictor selects the iteration count from \(\{3,4,5\}\), and a decoder fuses the final and shallow features to reconstruct \(\hat{\mathbf Y}\).

\subsection{Video Encoding and Degradation Condition Extraction}\label{video-encoding-and-degradation-condition-extraction}

Given a degraded clip \(\mathbf X\), the video encoder \(E\) downsamples only spatial dimensions to obtain latent features \(\mathbf h_0\), while the shallow branch \(S\) retains full-resolution features \(\mathbf h_s\). We then combine \(\mathbf h_0\) with input channel statistics \(\mu(\mathbf X)\) to extract a degradation representation \(\mathbf z\) and a severity-related scalar \(s\):

\[
\begin{aligned}
\mathbf h_0&=E(\mathbf X),\qquad \mathbf h_s=S(\mathbf X),\\
\mathbf z&=G_z\!\left([P(\mathbf h_0);\mu(\mathbf X)]\right),\qquad
s=\sigma(\mathbf w_s^\top\mathbf z+b_s),
\end{aligned}
\tag{1}
\]

Here \(P\) globally averages spatio-temporal features, \(G_z\) maps degradation conditions, and \(\sigma\) is Sigmoid; \(\mathbf z\) and \(s\) provide clip-level conditions for later iterations.

\subsection{Degradation-Conditioned Low-Rank Loop Embedding}\label{degradation-conditioned-low-rank-loop-embedding}

The degradation representation \(\mathbf z\) and scalar \(s\) describe restoration needs but do not distinguish loop stages. We therefore combine them with the sinusoidal position encoding~\citep{NIPS2017_3f5ee243} \(\mathbf p_\tau\) of iteration \(\tau\), correcting the base iteration embedding through a low-rank residual. First, we generate degradation and stage encodings \(\mathbf g_{\mathrm{deg}}\) and \(\mathbf g_\tau\):

\[
\begin{aligned}
\mathbf g_{\mathrm{deg}}&=\tanh\!\left(\mathbf A[\mathbf z;s]+\mathbf a\right),\\
\mathbf g_\tau&=\tanh\!\left(\mathbf B\mathbf p_\tau+\mathbf b\right),
\end{aligned}
\tag{2}
\]

where \([\mathbf z;s]\) denotes concatenation, \(\mathbf A,\mathbf B\) are projection matrices, and \(\mathbf a,\mathbf b\) are biases. Both encodings lie in \(\mathbb R^r\): \(\mathbf g_{\mathrm{deg}}\) remains constant within a clip, while \(\mathbf g_\tau\) varies across stages.

Low-rank element-wise interactions have a precedent in bilinear pooling~\citep{kim2017hadamard}. Here, their element-wise interaction generates a conditional residual \(\Delta\mathbf e_\tau\), which is added to the base embedding \(G_p(\mathbf p_\tau)\) to obtain the current loop condition \(\mathbf e_\tau\):

\[
\begin{aligned}
\Delta\mathbf e_\tau&=\mathbf U\!\left(\mathbf g_{\mathrm{deg}}\odot\mathbf g_\tau\right),\\
\mathbf e_\tau&=G_p(\mathbf p_\tau)+\Delta\mathbf e_\tau,
\end{aligned}
\tag{3}
\]

where \(\odot\) denotes element-wise multiplication, and \(\mathbf U\in\mathbb R^{d_e\times r}\) maps the interaction into the embedding space. Choosing \(r<d_e\) restricts the correction to a low-dimensional subspace, limiting additional parameters.

This interaction makes degradation corrections stage-dependent, providing conditions for different inputs and iterations. Zero-initializing \(\mathbf U\) preserves the initial base embedding, while training learns degradation-dependent residuals. The next module combines \(\mathbf e_\tau\) with video state and loop history for position-wise modulation.

\subsection{Spatio-Temporal Iterative Adaptive Normalization}\label{spatio-temporal-iterative-adaptive-normalization}\label{momentum-context}\label{adaptive-modulation}\label{spatio-temporal-attention-update}

To adapt loop conditions to the current restoration state, IterAda3DN updates clip-level momentum \(\mathbf m_\tau\) using the iteration input \(\mathbf h_{\tau-1}\), loop embedding \(\mathbf e_\tau\), and historical state:

\[
\begin{aligned}
\mathbf m_0&=G_0\!\left([P(\mathbf h_0);\mathbf z]\right),\\
\mathbf m_\tau&=0.7\,G_m\!\left([P(\mathbf h_{\tau-1});\mathbf e_\tau;\mathbf m_{\tau-1}]\right)+0.3\,\mathbf m_{\tau-1},
\end{aligned}
\tag{4}\label{eq:momentum-update}
\]

where \(G_0,G_m\) initialize and update momentum. The state \(\mathbf m_\tau\) propagates cross-iteration context and is reinitialized for each clip.

Using \(\mathbf m_\tau\) and current features, we generate channel and spatio-temporal factors \(\mathbf M_c,\mathbf M_{st}\), combining them into a modulation mask \(\mathbf M_\tau\):

\[
\begin{aligned}
\mathbf M_c&=2\sigma(\mathbf W_c\mathbf m_\tau+\mathbf b_c),\\
\mathbf M_{st}&=2\sigma\!\left(\mathbf W_{st}\mathbf h_{\tau-1}+\operatorname{br}(\mathbf w_b^\top\mathbf m_\tau+b_b)\right),\\
\mathbf M_\tau&=\mathbf M_c\odot\mathbf M_{st},
\end{aligned}
\tag{5}
\]

where \(\operatorname{br}\) denotes broadcasting. Zero-initialized projections yield an initial mask of 1, which each LoopLayer within the iteration applies to its current features.

Conditional affine modulation has a precedent in FiLM~\citep{perez2018film}; here both local spatio-temporal features and global video context determine the parameters. For branch input \(\mathbf v\), we combine its global statistics with loop information as \(\mathbf c=[P(\mathbf v);\mathbf e_\tau;\mathbf m_\tau]\), then incorporate local features to generate scales \(\boldsymbol\gamma_b\) and shifts \(\boldsymbol\beta_b\):

\[
\begin{aligned}
f_b(\mathbf v,\mathbf c)&=\operatorname{GELU}\!\left(\mathbf W_{\mathrm{loc}}^{(b)}\mathbf v+\operatorname{br}(\mathbf W_{\mathrm{glob}}^{(b)}\mathbf c)\right),\\
[\boldsymbol\gamma_b;\boldsymbol\beta_b]&=\tanh\!\left(\mathbf W_{\mathrm{aff}}^{(b)}f_b(\mathbf v,\mathbf c)\right),
\end{aligned}
\tag{6}
\]

where \(b=1,2\) denote the attention and expert branches. Local mappings retain spatio-temporal variation, while global mappings introduce video state and cross-iteration context.

The generated \(\boldsymbol\gamma_b,\boldsymbol\beta_b\) apply position-wise channel normalization and affine modulation to \(\mathbf v\):

\[
\mathcal N(\mathbf v;\boldsymbol\gamma_b,\boldsymbol\beta_b)
=(1+\boldsymbol\gamma_b)\odot
\frac{\mathbf v}{\sqrt{\frac{1}{d}\sum_{j=1}^{d}v_j^2+\varepsilon}}+\boldsymbol\beta_b,
\tag{7}
\]

where \(v_j\) is the \(j\)-th channel value at the current position, \(d\) is the channel count, and \(\varepsilon\) ensures stability. Normalization uses the channel root mean square~\citep{NEURIPS2019_1e8a1942}, with affine parameters varying across spatio-temporal positions.

For current layer features \(\mathbf h\), we first apply the mask to obtain \(\tilde{\mathbf h}=\mathbf M_\tau\odot\mathbf h\), then produce an updated state \(\mathbf u\) through the attention branch:

\[
\mathbf u=\mathbf h+\boldsymbol\alpha\odot
R_1\!\left(\mathcal A\!\left(\mathcal N(\tilde{\mathbf h};\boldsymbol\gamma_1,\boldsymbol\beta_1)\right)\right),
\tag{8}
\]

where \(\boldsymbol\alpha\) is a zero-initialized position-wise gate generated from modulation features, and \(R_1\) is channel normalization with learnable scaling. The operator \(\mathcal A\) combines frame-level temporal attention with local spatio-temporal convolution.

\subsection{State-Aware Expert Modulation}\label{state-aware-expert-modulation}\label{expert-branch-conditions}\label{sparse-expert-routing}\label{expert-feature-update}

Using the attention-updated state \(\mathbf u\), loop embedding, and momentum, the expert branch forms \(\mathbf c_2\) and obtains expert input \(\mathbf v_2\) through affine modulation:

\[
\begin{gathered}
\mathbf c_2=[P(\mathbf u);\mathbf e_\tau;\mathbf m_\tau],\\
(\boldsymbol\gamma_2,\boldsymbol\beta_2)=G_2(\mathbf u,\mathbf c_2),\\
\mathbf v_2=\mathcal N(\mathbf u;\boldsymbol\gamma_2,\boldsymbol\beta_2),
\end{gathered}
\tag{9}
\]

Here \(G_2\) is independent of the attention branch, so expert modulation uses the updated state.

Pooled features \(P(\mathbf v_2)\) and momentum \(\mathbf m_\tau\) yield routing scores \(\mathbf q\) that select \(k_e\) experts in \(\mathcal S\):

\[
\begin{aligned}
\mathbf q&=\operatorname{nrm}(\mathbf W_xP(\mathbf v_2))
+\tanh(\rho)\operatorname{nrm}(\mathbf W_m\mathbf m_\tau),\\
\mathcal S&=\operatorname{TopK}_{k_e}(\mathbf q),\qquad
\pi_i=\frac{\exp(q_i)}{\sum_{j\in\mathcal S}\exp(q_j)},
\end{aligned}
\tag{10}
\]

where \(\operatorname{nrm}\) is root mean square normalization, \(\rho\) is a scalar gate, and \(\pi_i\) is the normalized weight of a selected expert. This top-$k_e$ selection follows sparse expert routing~\citep{shazeer2017}. All positions within a clip use the same expert set.

Shared experts have a precedent in DeepSeekMoE~\citep{dai-etal-2024-deepseekmoe}. Outputs from routed experts \(E_i\) and shared experts are aggregated as \(\mathcal E\) to update the features:

\[
\begin{aligned}
\mathcal E(\mathbf v_2,\mathbf m_\tau)
&=\sum_{i\in\mathcal S}\pi_iE_i(\mathbf v_2)
+\frac{1}{N_s}\sum_{j=1}^{N_s}E_j^{\mathrm{shared}}(\mathbf v_2),\\
\mathbf h_{\mathrm{next}}&=\mathbf u+R_2\!\left(\mathcal E(\mathbf v_2,\mathbf m_\tau)\right),
\end{aligned}
\tag{11}
\]

where \(E_i\) comprises temporal convolution, spatial convolution, and pointwise projection, \(N_s\) counts shared experts, and \(R_2\) is channel normalization with learnable scaling. Each iteration executes two LoopLayers, each containing 16 routed and 2 shared experts and activating 4 routed experts. The iteration output \(\mathbf h_\tau\) proceeds to further refinement or reconstruction.

\subsection{Output Reconstruction and Dynamic Depth Scheduling}\label{output-reconstruction-and-dynamic-depth-scheduling}\label{residual-reconstruction}\label{dynamic-depth-scheduling}

After \(K\) refinement iterations, the decoder \(D\) restores \(\mathbf h_K\) to the input resolution. The residual fusion network \(F_{\mathrm{fuse}}\) combines it with shallow features \(\mathbf h_s\) to reconstruct \(\hat{\mathbf Y}\):

\[
\hat{\mathbf Y}=\operatorname{clip}_{[0,1]}\!\left(
\mathbf X+F_{\mathrm{fuse}}\!\left([D(\mathbf h_K);\mathbf h_s]\right)\right),
\tag{12}
\]

where features are cropped to the original size before fusion. The output preserves the input temporal length and spatial resolution and is clipped to \([0,1]\).

To adapt computation to the input, the depth predictor maps \([\mathbf z;s]\) to probabilities \(p_4,p_5\) of requiring at least four or five iterations, progressively determining \(K\):

\[
K=3+\mathbb I[p_4\geq0.5]+\mathbb I[p_4\geq0.5\land p_5\geq0.5],
\tag{13}
\]

where \(\mathbb I\) is the indicator function, and \(p_4,p_5\) are Sigmoid outputs of predicted logits. This selects \(K\in\{3,4,5\}\), permitting five iterations only when the four-iteration condition holds.

\subsection{Loss Functions and Depth Supervision}\label{loss-functions}\label{video-restoration-loss}\label{depth-label-generation}\label{depth-prediction-loss}

Training proceeds in two stages. The first stage freezes the depth predictor and optimizes restoration at four iterations; the second freezes the restoration network and condition extractor to train the depth predictor. The stage-wise total loss is

\[
\mathcal L_{\mathrm{total}}=
\begin{cases}
\mathcal L_{\mathrm{rec}}^{(4)}+\mathcal L_{\mathrm{aux}}^{(4)}, & \text{Stage 1},\\
\mathcal L_K, & \text{Stage 2}.
\end{cases}
\tag{14}\label{eq:total-loss}
\]

The restoration loss combines pixel, temporal-difference, and frequency terms; the auxiliary loss constrains modulation and expert routing. The ordinal loss \(\mathcal L_K\) supervises depth prediction. Appendix~\ref{app:training-objective} defines these terms and the quality-cost criterion used to construct depth labels.

\section{Experiments}\label{experiments}

\subsection{Experimental Setup}\label{experimental-setup}

\textbf{Data and evaluation.} We use the official UHV-4K-AIO split~\citep{wang2026mocramixturecompositionalrank1} and one model for all four restoration tasks. Five-frame windows are evaluated on their center frames at native \(3840\times2160\) resolution. We report restoration quality and temporal consistency metrics with equal weighting across tasks; dataset splits, metric definitions, and evaluation protocols are detailed in Appendix~\ref{app:data}--\ref{app:metrics}.

\textbf{Implementation and comparisons.} Training uses AdamW~\citep{loshchilov2019decoupled} and FP16 mixed precision~\citep{micikevicius2018mixed} on one NVIDIA V100 (32GB), with aligned \(256\times256\) crops, a batch size of 2, gradient clipping, and cosine learning rate decay with warmup. The restoration network is first trained for 100 epochs with four iterations. The restoration network and condition extraction module are then frozen, depth labels are generated as detailed in Appendix~\ref{app:training-objective}, and the depth predictor is trained for another 100 epochs. We compare methods on the official UHV-4K-AIO test split using the same evaluation metrics.

\subsection{Main Results on UHV-4K-AIO}\label{main-results-on-uhv-4k-aio}

\begin{figure}[!b]
\centering
\includegraphics[width=.96\linewidth]{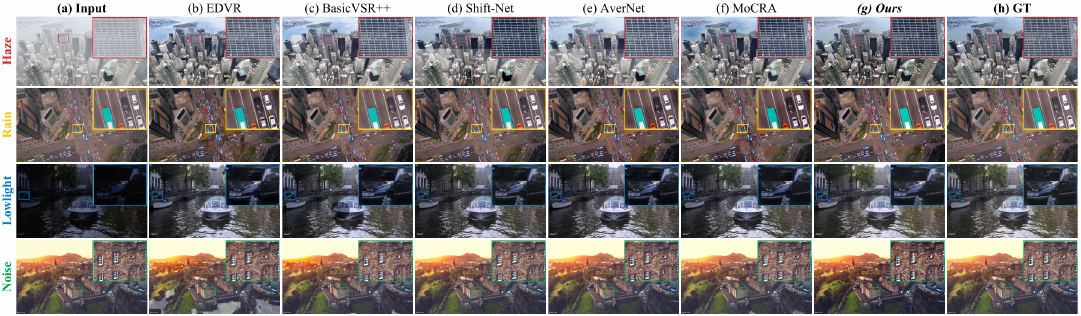}
\caption{Qualitative comparison of the top six video methods by mean PSNR on UHV-4K-AIO.}
\label{fig:qualitative}
\end{figure}

Table~\visualref{tab:main} shows that LoopMoEVR attains the highest mean PSNR with 0.884M parameters, exceeding MoCRA by 0.64 dB with roughly one-quarter as many parameters. A single configuration covers all four degradations and achieves the best PSNR and SSIM for dehazing and low-light enhancement. Figure~\visualref{fig:qualitative} compares identically cropped details from six video methods.

\begin{table}[!htbp]
\centering
\setlength{\tabcolsep}{1.1pt}
\renewcommand{\arraystretch}{1.3}
\fontsize{7}{8}\selectfont
\begin{adjustbox}{max width=\linewidth}
\begin{tabular}{@{}ll|rrr|rrr|rrr|rrr|rrr|rr|rr@{}}
\toprule
\multirow{2}{*}{Method} & \multirow{2}{*}{Venue} & \multicolumn{3}{c|}{Dehaze} & \multicolumn{3}{c|}{Derain} & \multicolumn{3}{c|}{Denoise} & \multicolumn{3}{c|}{Low light} & \multicolumn{3}{c|}{Mean} & \multicolumn{2}{c|}{No-reference} & \multicolumn{2}{c}{Efficiency at 4K} \\
\cmidrule(lr){3-5}\cmidrule(lr){6-8}\cmidrule(lr){9-11}\cmidrule(lr){12-14}\cmidrule(lr){15-17}\cmidrule(lr){18-19}\cmidrule(l){20-21}
 &  & PSNR$\uparrow$ & SSIM$\uparrow$ & LPIPS$\downarrow$ & PSNR$\uparrow$ & SSIM$\uparrow$ & LPIPS$\downarrow$ & PSNR$\uparrow$ & SSIM$\uparrow$ & LPIPS$\downarrow$ & PSNR$\uparrow$ & SSIM$\uparrow$ & LPIPS$\downarrow$ & PSNR$\uparrow$ & SSIM$\uparrow$ & LPIPS$\downarrow$ & MUSIQ$\uparrow$ & MANIQA$\uparrow$ & \shortstack{Params.\\(M)} & \shortstack{Reported\\time (s)} \\
\midrule
\textcolor{gray}{Degraded input} & \textcolor{gray}{--} & \textcolor{gray}{13.19} & \textcolor{gray}{0.769} & \textcolor{gray}{0.226} & \textcolor{gray}{27.24} & \textcolor{gray}{0.916} & \textcolor{gray}{0.131} & \textcolor{gray}{27.73} & \textcolor{gray}{0.558} & \textcolor{gray}{0.493} & \textcolor{gray}{9.06} & \textcolor{gray}{0.232} & \textcolor{gray}{0.966} & \textcolor{gray}{19.31} & \textcolor{gray}{0.619} & \textcolor{gray}{0.454} & \textcolor{gray}{31.14} & \textcolor{gray}{0.223} & \textcolor{gray}{--} & \textcolor{gray}{--} \\
\midrule
\rowcolor{black!8}\multicolumn{21}{c}{\textit{(a) Image methods}} \\
\midrule
AirNet & CVPR'22 & 22.04 & 0.939 & 0.051 & 28.99 & 0.973 & 0.022 & 37.54 & 0.968 & 0.036 & 19.86 & 0.826 & 0.221 & 27.11 & 0.927 & 0.083 & 36.04 & 0.230 & 8.93 & 46.69 \\
TransWeather & CVPR'22 & 19.39 & 0.883 & 0.105 & 27.94 & 0.918 & 0.129 & 35.15 & 0.940 & 0.110 & 20.98 & 0.765 & 0.347 & 25.87 & 0.877 & 0.173 & 33.50 & 0.222 & 38.05 & 1.72 \\
UHD-processer & CVPR'25 & 20.38 & 0.899 & 0.106 & 29.79 & 0.934 & 0.113 & 33.09 & 0.784 & 0.340 & 23.82 & 0.693 & 0.490 & 26.77 & 0.828 & 0.262 & 32.60 & 0.210 & 1.71 & 4.07 \\
DRNet & TMM'26 & 14.16 & 0.812 & 0.183 & 26.34 & 0.916 & 0.127 & 24.29 & 0.806 & 0.270 & 9.67 & 0.369 & 0.686 & 18.62 & 0.726 & 0.317 & 33.06 & 0.212 & 16.49 & 18.93 \\
R2R & CVPR'26 & 11.74 & 0.752 & 0.264 & 16.49 & 0.854 & 0.163 & 25.68 & 0.880 & 0.177 & 9.08 & 0.332 & 0.676 & 15.75 & 0.705 & 0.320 & 31.34 & 0.219 & 27.58 & 1.74 \\
\midrule
\rowcolor{black!8}\multicolumn{21}{c}{\textit{(b) Video methods}} \\
\midrule
EDVR & CVPRW'19 & 21.08 & 0.921 & 0.093 & 23.40 & 0.937 & 0.081 & 24.11 & 0.636 & 0.272 & 19.25 & 0.533 & 0.412 & 21.96 & 0.757 & 0.214 & 34.12 & 0.212 & 2.85 & 21.19 \\
BasicVSR++ & CVPR'22 & 21.78 & \underline{0.932} & 0.061 & 26.31 & \underline{0.955} & \underline{0.059} & 34.55 & \underline{0.948} & \underline{0.062} & 22.34 & 0.819 & 0.246 & 26.25 & 0.914 & \underline{0.107} & \underline{35.81} & 0.218 & 6.45 & 43.93 \\
RVRT & NeurIPS'22 & 19.77 & 0.827 & 0.191 & 23.63 & 0.899 & 0.148 & 19.83 & 0.842 & 0.189 & 17.29 & 0.671 & 0.449 & 20.13 & 0.810 & 0.244 & 35.79 & 0.212 & 3.06 & 76.62 \\
Shift-Net & CVPR'23 & 22.24 & 0.924 & 0.054 & 27.28 & 0.919 & 0.128 & \underline{36.19} & 0.917 & 0.142 & 21.76 & 0.765 & 0.352 & 26.87 & 0.881 & 0.169 & 34.22 & 0.219 & \underline{2.17} & 9.22 \\
ViWS-Net & ICCV'23 & 22.23 & 0.913 & 0.068 & 23.59 & 0.907 & 0.096 & 28.57 & 0.894 & 0.154 & 8.44 & 0.279 & 0.573 & 20.71 & 0.748 & 0.223 & 33.08 & 0.221 & 57.67 & 8.91 \\
AverNet & NeurIPS'24 & 21.25 & 0.922 & 0.063 & 27.58 & \textbf{0.957} & \textbf{0.039} & \textbf{37.32} & \textbf{0.957} & \textbf{0.050} & 21.54 & 0.830 & \textbf{0.222} & 26.92 & \underline{0.917} & \textbf{0.094} & \textbf{36.07} & \underline{0.225} & 13.72 & 51.63 \\
MoCRA & arxiv'2026 & \underline{23.47} & \underline{0.932} & \underline{0.050} & \textbf{28.70} & 0.931 & 0.095 & 32.07 & 0.907 & 0.165 & \underline{24.66} & \underline{0.834} & 0.252 & \underline{27.23} & 0.901 & 0.141 & 34.40 & 0.224 & 3.64 & \textbf{0.48} \\
\midrule
\rowcolor{black!5}
\textbf{LoopMoEVR (Ours)} & - & \textbf{24.12} & \textbf{0.941} & \textbf{0.046} & \underline{28.20} & 0.948 & 0.081 & 33.85 & 0.933 & 0.138 & \textbf{25.30} & \textbf{0.849} & \underline{0.231} & \textbf{27.87} & \textbf{0.918} & 0.124 & 35.62 & \textbf{0.226} & \textbf{0.884} & \underline{0.51} \\
\bottomrule
\end{tabular}
\end{adjustbox}
\normalsize
\caption{UHV-4K-AIO restoration results at native 4K.}
\label{tab:main}
\end{table}

\subsection{Robustness to Time-Varying Degradations}\label{robustness-to-time-varying-degradations}

To evaluate robustness to within-clip degradation changes, we switch degradation types every \(\Delta t\in\{6,3\}\) frames in UHV-4K-AIO test clips. LoopMoEVR achieves 27.68 and 27.85 dB mean PSNR, respectively, at most 0.19 dB below the static setting, indicating robustness to within-clip degradation changes; complete results appear in Appendix~\ref{app:dynamic-temporal}.

\subsection{Temporal Stability}\label{temporal-stability}

LoopMoEVR obtains a tLPIPS of 0.1182 and a Warping Error of 0.0297, compared with AverNet's 0.1238 and 0.0296, respectively. The full comparison appears in Appendix~\ref{app:dynamic-temporal}. These results demonstrate that the compact model maintains competitive inter-frame consistency at both levels.

Temporal difference supervision constrains inter-frame changes during training, while video-state modulation incorporates within-clip spatio-temporal information during restoration. These mechanisms address temporal consistency through the optimization objective and feature updates, respectively. Together with the main results, the evaluation shows that LoopMoEVR improves average restoration quality while maintaining competitive temporal stability.

\subsection{Analysis of Non-UHD Resolution Results}\label{non-uhd-results-analysis}

LoopMoEVR remains effective at non-UHD resolutions but shows a smaller advantage than in UHD restoration (Appendix~\ref{app:non-uhd-results}). Designed for UHD inputs, its looped refinement exploits rich structural and degradation cues; fewer high-frequency details at lower resolutions limit iterative gains.

\subsection{Real-World UHD Image Restoration}\label{real-world-uhd-restoration}

We apply LoopMoEVR trained on UHV-4K-AIO directly to real-world UHD images. As shown in Appendix~\ref{app:real-world-uhd}, it suppresses real-world degradations while preserving clear local structures and texture details, achieving the best visual restoration quality among the compared methods.

\subsection{Ablation Studies}\label{ablation-experiments}

\textbf{Core components.} The full model exceeds the LoopMoE-style video baseline~\citep{chen2026loopmoeunifyingiterativecomputation} by 1.20 dB PSNR (Table~\visualref{tab:ablation-summary}(a)). All listed components improve restoration; momentum also benefits temporal consistency.

\textbf{Loop-depth allocation.} Dynamic depth saves 8.9\% inference time over fixed four-step inference at similar quality and gains 0.51 dB PSNR over distribution-matched random allocation (Table~\visualref{tab:ablation-summary}(b)), demonstrating effective input-dependent computation.

\textbf{Loop-conditioning formulation.} Low-rank interaction exceeds the concatenation MLP by 0.38 dB PSNR and achieves the lowest tLPIPS (Table~\visualref{tab:ablation-summary}(c)), supporting the interaction of degradation information and iteration position. Visual comparisons appear in Appendix~\ref{app:ablation-vis}.

\begin{figure}[!htbp]
\centering
\noindent\ablationtile{.136\linewidth}{a}{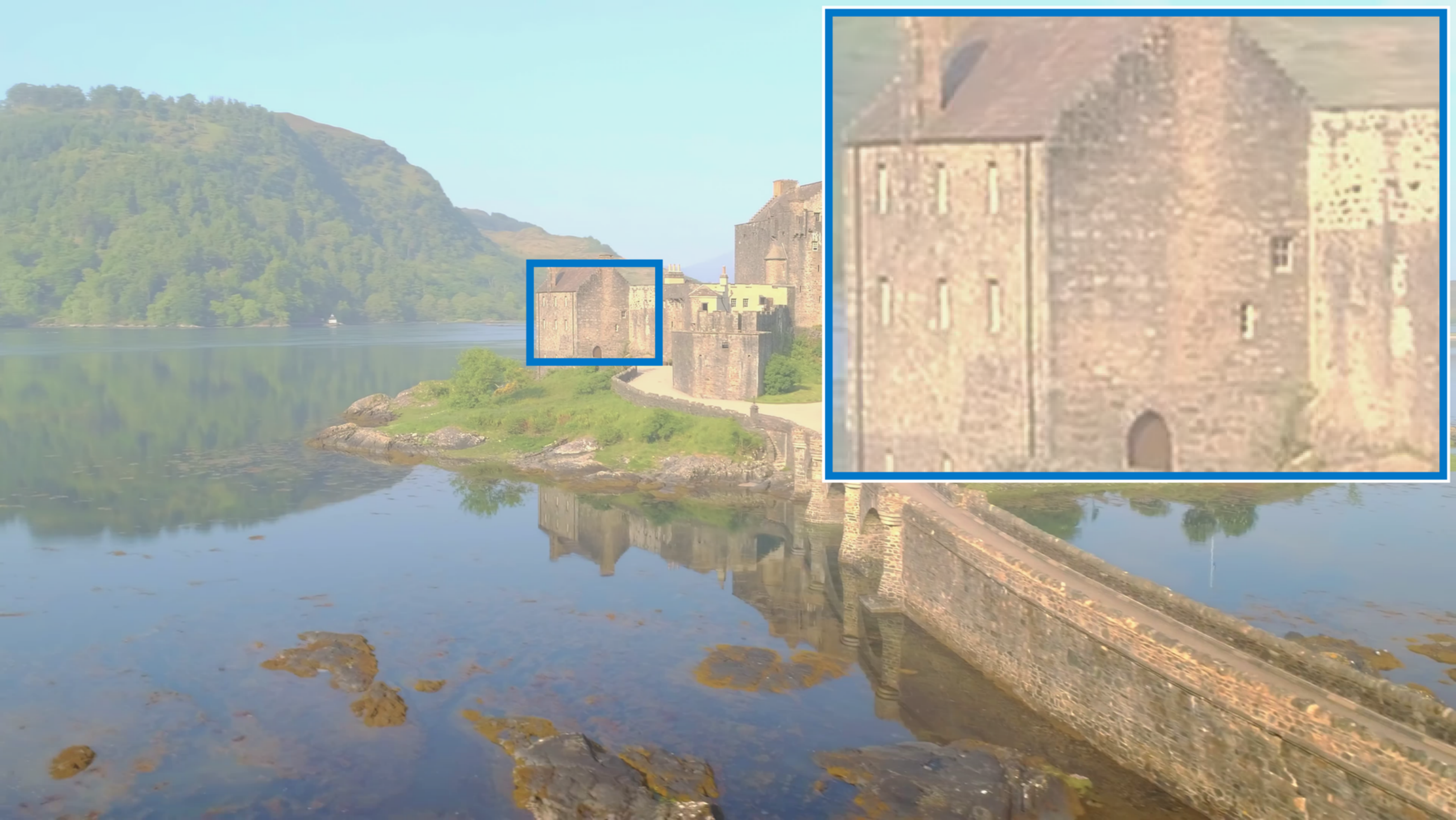}{Input}\hfill%
\ablationtile{.136\linewidth}{b}{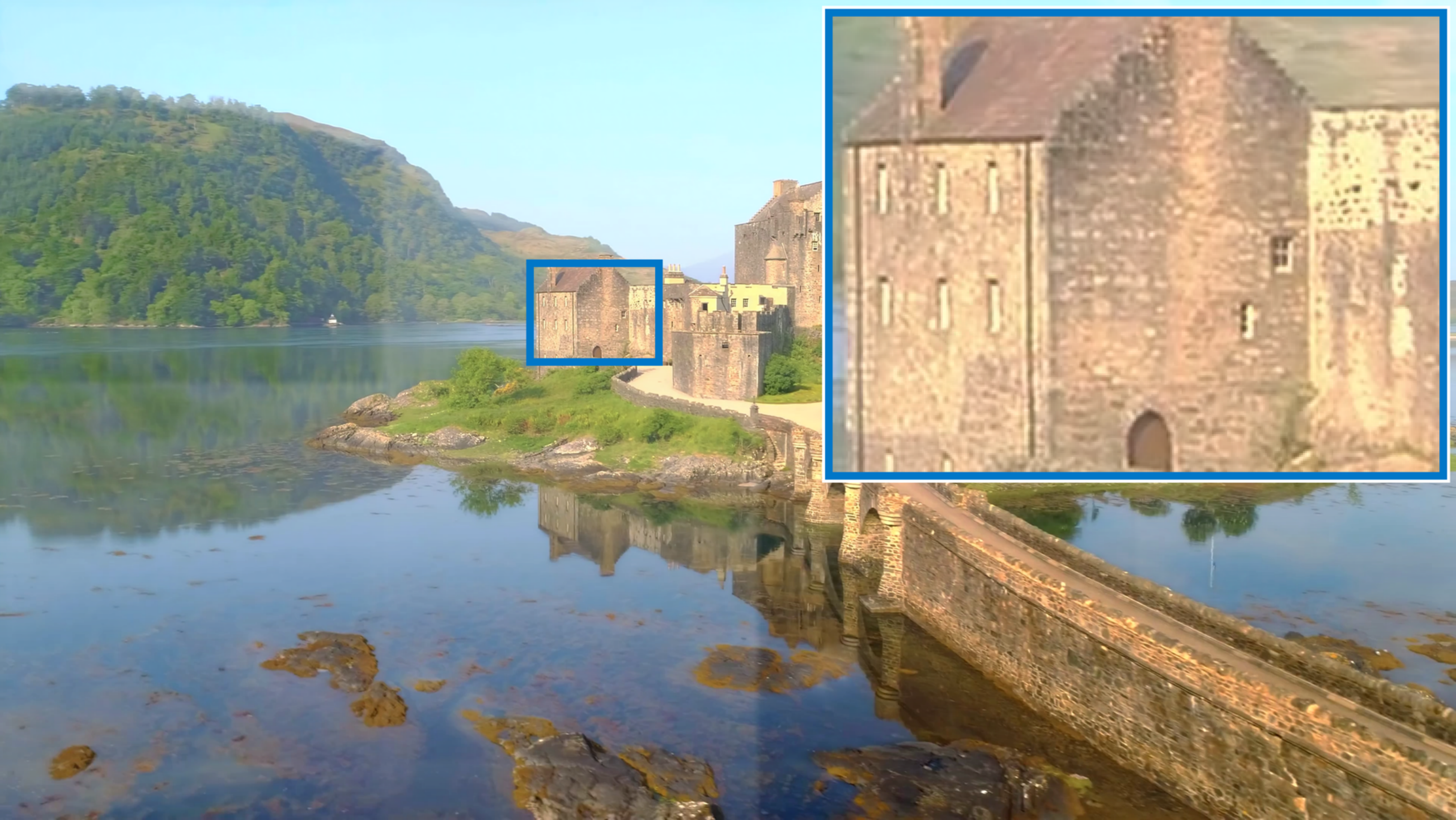}{\(\mathcal L_1\) only}\hfill%
\ablationtile{.136\linewidth}{c}{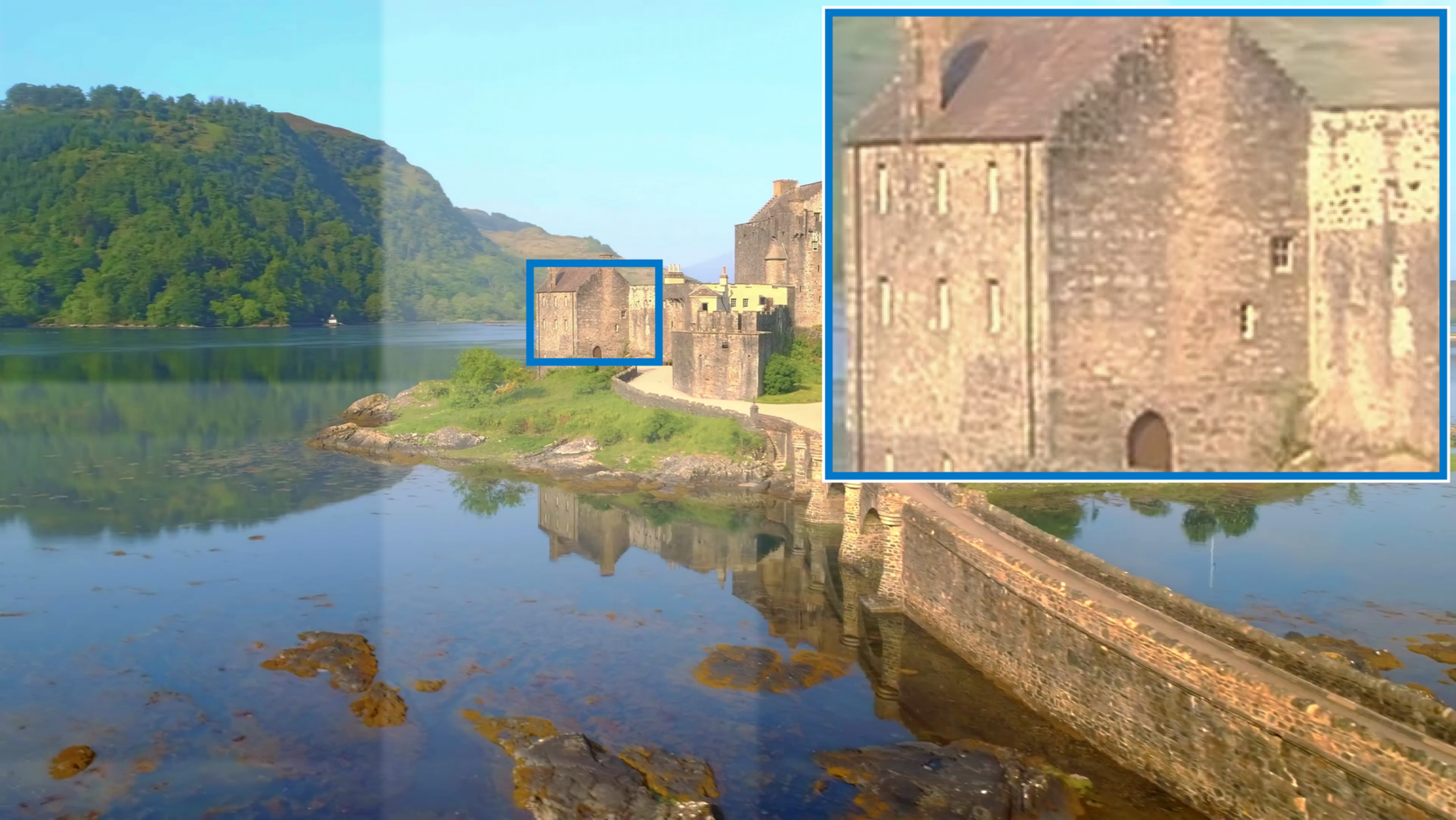}{w/o \(\mathcal L_t\)}\hfill%
\ablationtile{.136\linewidth}{d}{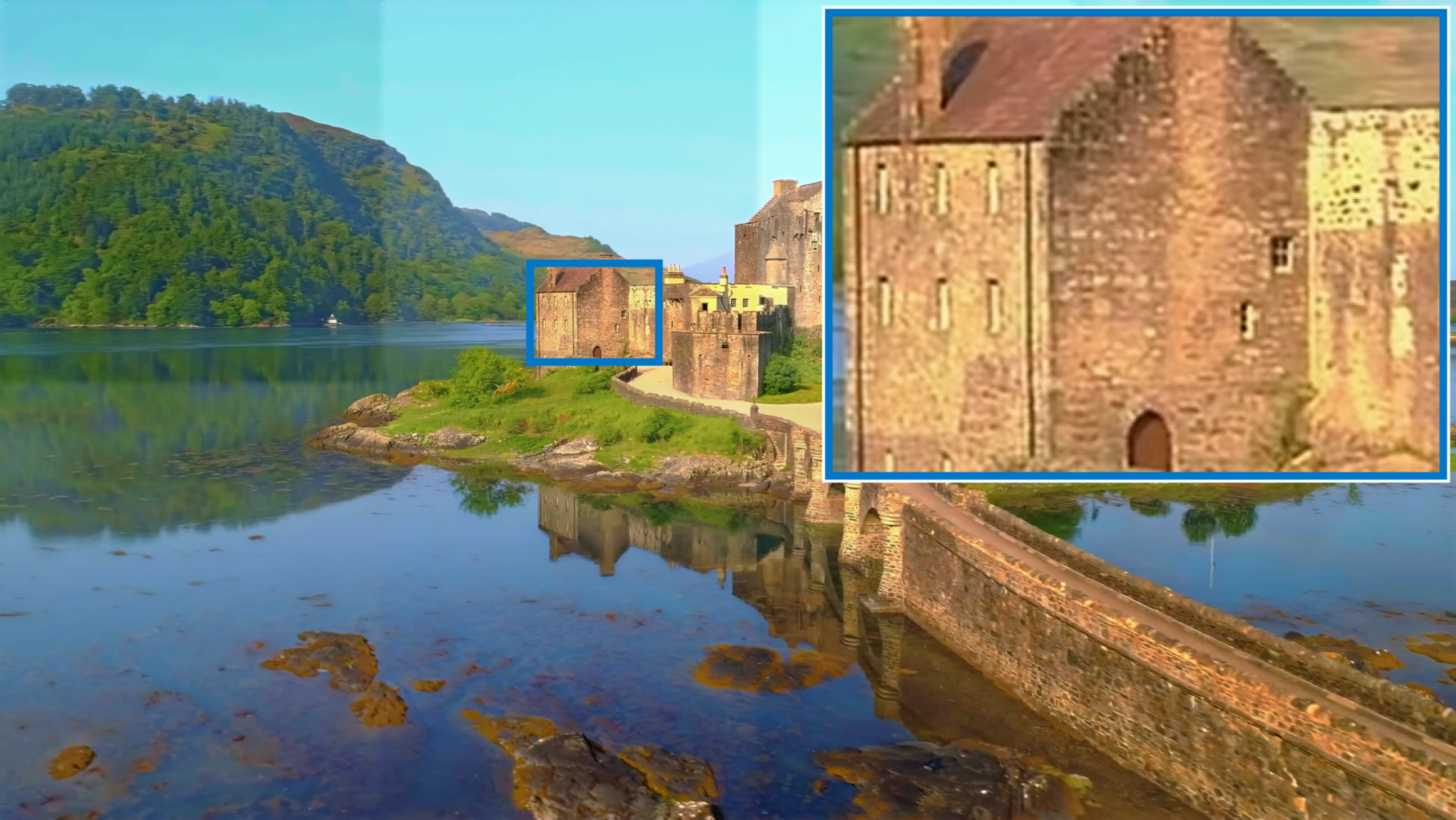}{w/o \(\mathcal L_f\)}\hfill%
\ablationtile{.136\linewidth}{e}{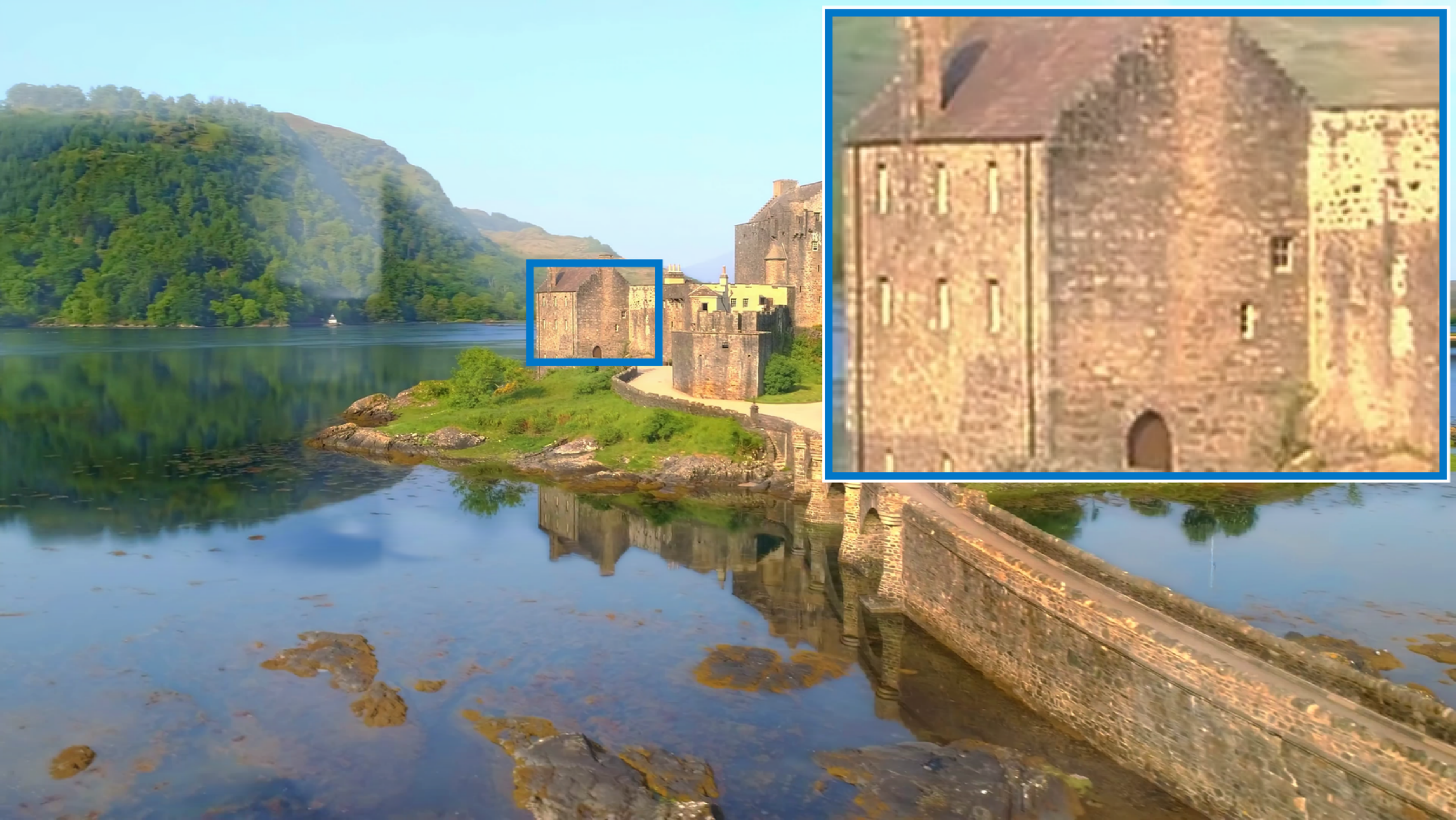}{w/o \(\mathcal L_{\mathrm{aux}}\)}\hfill%
\ablationtile{.136\linewidth}{f}{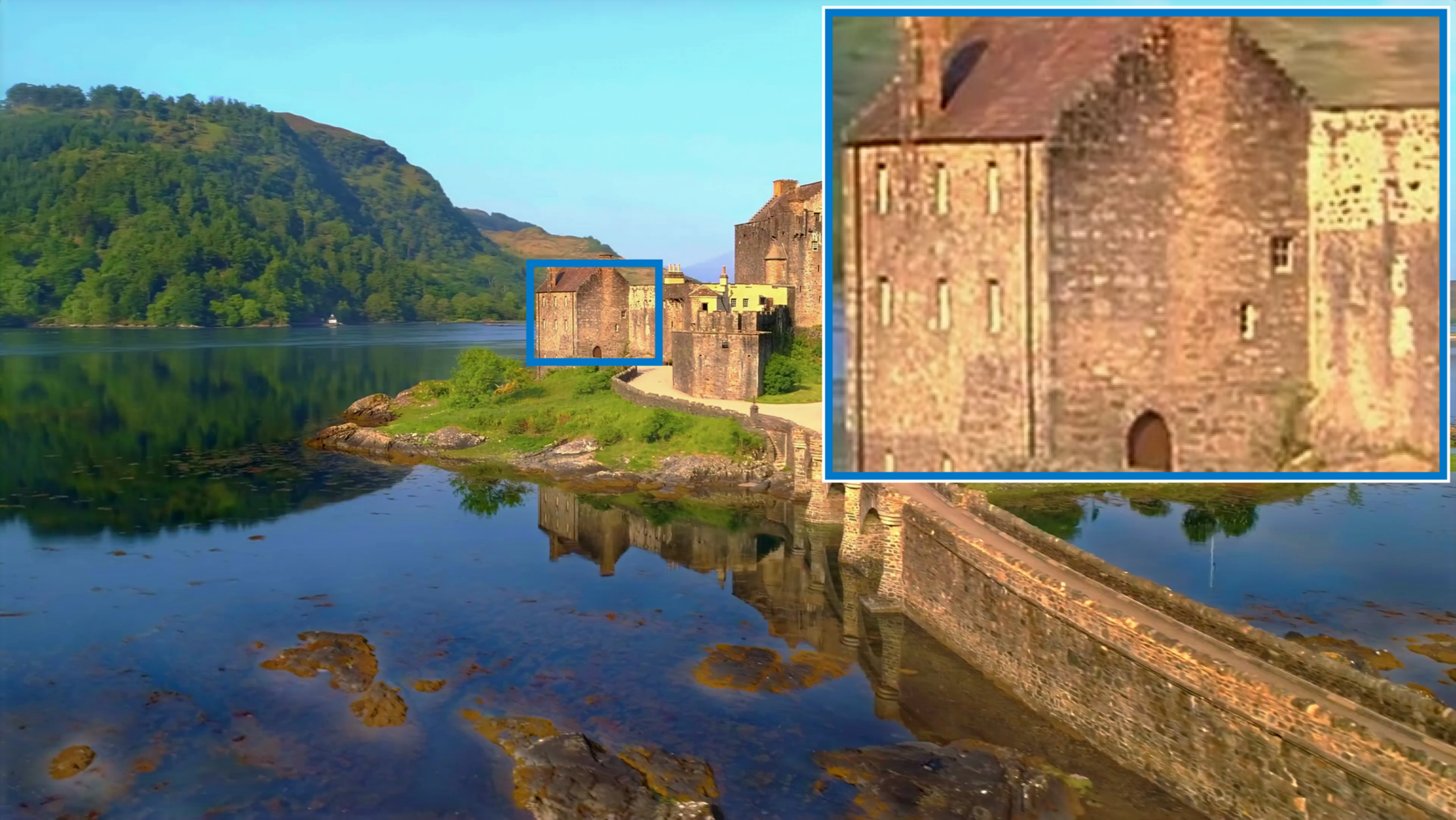}{\textbf{Ours}}\hfill%
\ablationtile{.136\linewidth}{g}{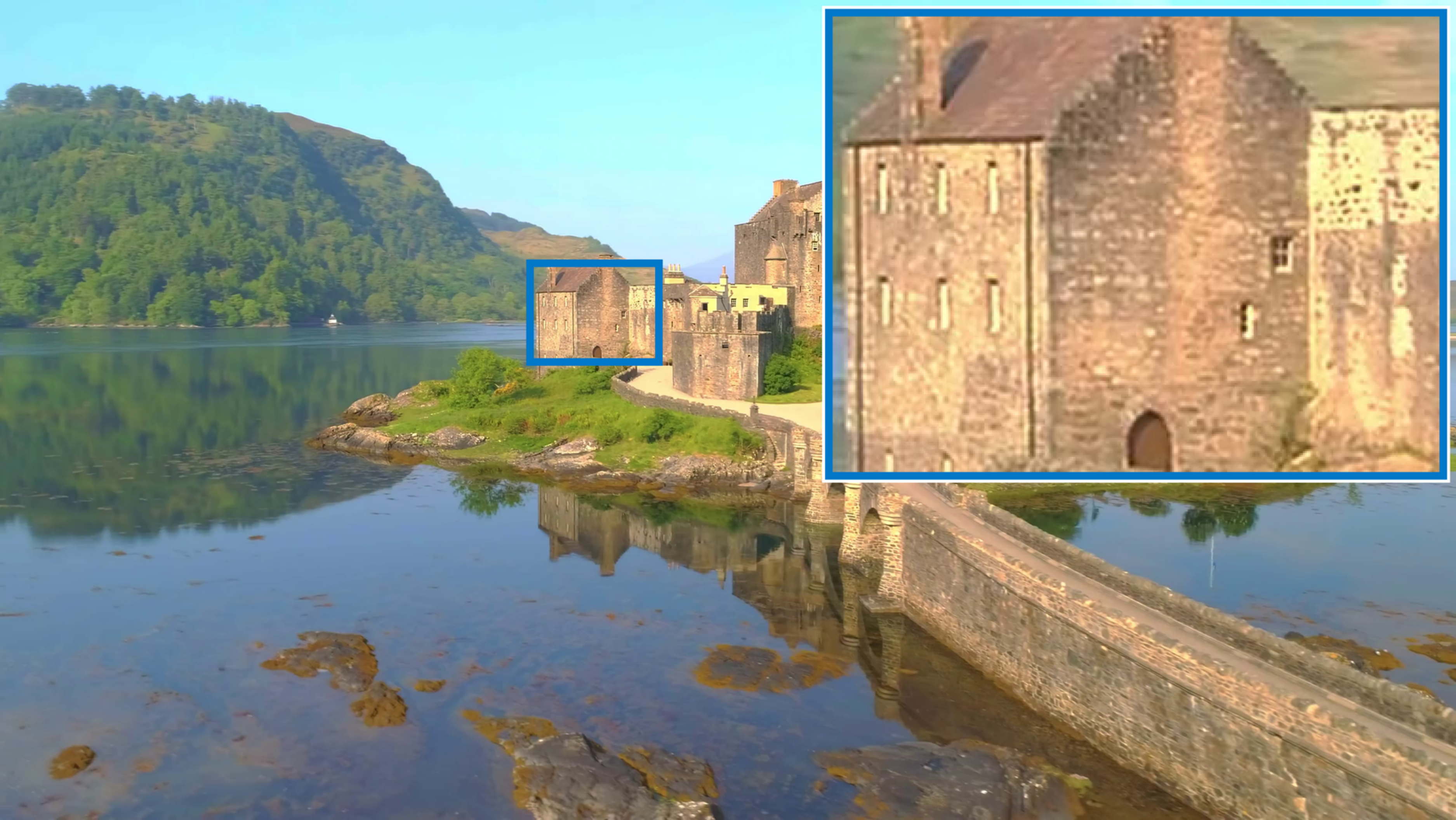}{GT}
\caption{Visual comparison of loss-function ablations.}
\label{fig:ablation-loss}
\end{figure}

\textbf{Loss functions.} The full objective yields the lowest LPIPS and tLPIPS (Table~\visualref{tab:ablation-summary}(d)); temporal supervision improves consistency, while frequency and auxiliary terms aid restoration. Figure~\visualref{fig:ablation-loss} presents visual comparisons.

\begin{table}[!t]
\centering
\begingroup
\setlength{\tabcolsep}{1.8pt}
\renewcommand{\arraystretch}{1.05}
\fontsize{7}{8}\selectfont
\begin{minipage}[t]{.49\linewidth}
\vspace{0pt}\textbf{(a) Core components}\par\vspace{2pt}
\begin{tabularx}{\linewidth}{@{}>{\raggedright\arraybackslash}Xrr@{}}
\toprule
Configuration & PSNR & tLPIPS \\
\midrule
LoopMoE-style baseline & 26.78 & 0.1614 \\
w/o degr. residual & 27.12 & 0.1424 \\
w/o dual-branch cond. & 27.00 & 0.1484 \\
w/o expert cond. & 27.37 & 0.1354 \\
w/o momentum state & 27.30 & 0.1434 \\
\midrule
\rowcolor{black!5}\textbf{Full model (Ours)} & \textbf{27.98} & \textbf{0.1144} \\
\bottomrule
\end{tabularx}
\end{minipage}\hfill
\begin{minipage}[t]{.49\linewidth}
\vspace{0pt}\textbf{(b) Loop-depth allocation}\par\vspace{2pt}
\begin{tabularx}{\linewidth}{@{}>{\raggedright\arraybackslash}Xrrrr@{}}
\toprule
Policy & Mean $K$ & PSNR & tLPIPS & ms \\
\midrule
Fixed $K=2$ & 2.00 & 26.35 & 0.1694 & \textbf{383} \\
Fixed $K=3$ & 3.00 & 27.24 & 0.1394 & 470 \\
Fixed $K=4$ & 4.00 & 27.98 & \textbf{0.1144} & 560 \\
Fixed $K=5$ & 5.00 & 28.08 & 0.1154 & 650 \\
Fixed $K=6$ & 6.00 & \textbf{28.11} & 0.1169 & 748 \\
Matched random & 3.45 & 27.36 & 0.1422 & 532 \\
\midrule
\rowcolor{black!5}\textbf{Dynamic (Ours)} & 3.45 & 27.87 & 0.1182 & 510 \\
\bottomrule
\end{tabularx}
\end{minipage}
\par\vspace{5pt}
\begin{minipage}[t]{.49\linewidth}
\vspace{0pt}\textbf{(c) Loop conditioning}\par\vspace{2pt}
\begin{tabularx}{\linewidth}{@{}>{\raggedright\arraybackslash}Xrrr@{}}
\toprule
Form & \shortstack{Params\\(k)} & PSNR & tLPIPS \\
\midrule
Base iteration encoding & 8.0 & 27.12 & 0.1424 \\
Additive conditioning & 14.0 & 27.41 & 0.1324 \\
Concatenation MLP & 14.0 & 27.60 & 0.1254 \\
\midrule
\rowcolor{black!5}\textbf{Low-rank (Ours)} & 14.0 & \textbf{27.98} & \textbf{0.1144} \\
\bottomrule
\end{tabularx}
\end{minipage}\hfill
\begin{minipage}[t]{.49\linewidth}
\vspace{0pt}\textbf{(d) Loss functions}\par\vspace{2pt}
\begin{tabularx}{\linewidth}{@{}>{\raggedright\arraybackslash}Xrrr@{}}
\toprule
Configuration & PSNR & LPIPS & tLPIPS \\
\midrule
$\mathcal L_1$ only & 26.96 & 0.159 & 0.1684 \\
w/o $\mathcal L_t$ & \textbf{28.06} & 0.127 & 0.1524 \\
w/o $\mathcal L_f$ & 27.49 & 0.145 & 0.1234 \\
w/o $\mathcal L_{\mathrm{aux}}$ & 27.40 & 0.143 & 0.1364 \\
\midrule
\rowcolor{black!5}\textbf{Full loss (Ours)} & 27.98 & \textbf{0.122} & \textbf{0.1144} \\
\bottomrule
\end{tabularx}
\end{minipage}
\endgroup
\caption{LoopMoEVR ablations. Panels (a), (c), and (d) use four-iteration models; panel (b) uses one checkpoint trained at four iterations.}
\label{tab:ablation-summary}
\end{table}

\section{Discussion: Why Does Looped Computation Benefit Unified UHD Video Restoration?}\label{discussion-looped-uhd}

Unified 4K restoration must address haze and low light visible at reduced resolution, alongside rain and noise requiring native-scale information~\citep{wang2026mocramixturecompositionalrank1,Liu_2025_CVPR}. Backbone reuse increases processing depth with shared parameters~\citep{dehghani2018universal}:
\[
\mathbf h_\tau=F_\theta(\mathbf h_{\tau-1};\mathbf e_\tau,\mathbf m_\tau),
\qquad
P_{\mathrm{shared}}\approx P_\theta,\quad
P_{\mathrm{independent}}\approx KP_\theta,
\]
Here \(F_\theta\) updates video features conditioned on loop embedding \(\mathbf e_\tau\) and momentum state \(\mathbf m_\tau\). Its \(P_\theta\) backbone parameters are reused over \(K\) iterations, whereas independent backbones require approximately \(KP_\theta\) parameters. LoopMoEVR refines compact spatio-temporal features and fuses native-resolution shallow features for 4K reconstruction.

\begin{figure}[!htbp]
\centering
\input{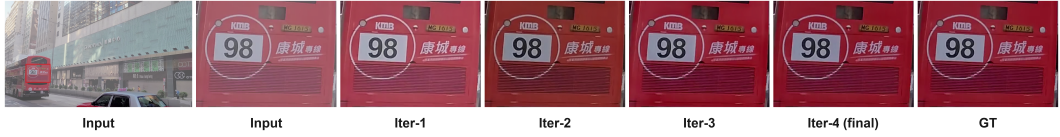}
\caption{Iterative dehazing results of LoopMoEVR.}
\label{fig:discussion-iterations}
\end{figure}

Degradation, video-state, and iteration conditions adapt shared-backbone updates to the restoration state; prior work supports the role of timestep encoding~\citep{pmlr-v267-xu25x}. Figure~\visualref{fig:discussion-iterations} illustrates progressive correction, and Appendix~\ref{app:condition-order} shows that the normal condition order outperforms reversed, shuffled, and repeated alternatives. The parameter-sharing comparison appears in Appendix~\ref{app:parameter-sharing}. LoopMoEVR combines state-dependent refinement with native-resolution reconstruction to enable unified 4K restoration with a compact model.

\FloatBarrier

\Needspace{9\baselineskip}
\section{Conclusion}\label{conclusion-and-future-work}

We propose LoopMoEVR, a unified UHD video restoration model based on the loop learning paradigm. It combines degradation-conditioned low-rank loop embeddings, IterAda3DN, and state-aware expert modulation to enable unified UHD video restoration with adaptive depth. With approximately 0.884M parameters, the model achieves the highest mean PSNR on UHV-4K-AIO while maintaining strong performance under time-varying degradations and in temporal-consistency evaluation. The mobile demonstration indicates its potential for resource-constrained deployment.

\clearpage
\section*{AI Use Statement}

Generative AI tools were used under the authors’ direction to assist with the development and refinement of hypotheses, experimental design, software implementation, literature search, result interpretation, and the drafting and editing of the manuscript. They were also used to assist with mathematical derivations and proof writing. All AI-assisted code, analyses, proofs, figures, and text were reviewed and verified by the authors. The authors take full responsibility for the final content of this work.

\bibliographystyle{loopmoevr}
\bibliography{references}

\clearpage
\appendix
\section*{Appendix Contents}
\noindent The supplementary material is organized into four parts: Section~\ref{app:architecture} specifies the model and training objective; Section~\ref{app:protocol} documents the evaluation protocol; Section~\ref{app:extended} reports quantitative ablations; and Section~\ref{app:qualitative} discusses supplementary visual evidence.
\par\smallskip
\noindent A. Architecture and Training Details \dotfill \pageref{app:architecture}\par
\noindent B. Experimental Protocol and Reproducibility \dotfill \pageref{app:protocol}\par
\noindent C. Extended Ablations \dotfill \pageref{app:extended}\par
\noindent D. Qualitative Evidence and Mechanism Analysis \dotfill \pageref{app:qualitative}
\par\bigskip

\section{Architecture and Training Details}\label{app:architecture}

\subsection{Forward Pass and Module Interfaces}\label{app:forward}
For each five-frame input window, the encoder produces spatially downsampled latent features and the shallow branch retains full-resolution detail. Global pooling of the latent features and input channel statistics provides the degradation representation and severity-related scalar. Their low-rank interaction with the iteration encoding yields the stage condition. At each iteration, clip-level momentum and the current features determine a modulation mask; two parameter-shared LoopLayers then update the spatio-temporal representation. Each LoopLayer has 16 routed and 2 shared experts, with 4 routed experts activated for a clip. The attention branch precedes the expert branch, so expert modulation uses the attention-updated local state. The decoder fuses the final latent state with shallow features and predicts a residual at the input resolution. The iteration count is selected from \(\{3,4,5\}\) by the depth predictor.

\subsection{Two-Stage Training Objective}\label{app:training-objective}

\begingroup
\setlength{\abovedisplayskip}{5pt plus 1pt minus 1pt}
\setlength{\belowdisplayskip}{5pt plus 1pt minus 1pt}
\setlength{\abovedisplayshortskip}{4pt plus 1pt minus 1pt}
\setlength{\belowdisplayshortskip}{4pt plus 1pt minus 1pt}

Training separates restoration learning from depth prediction. Given the predicted video \(\hat{\mathbf Y}\) and reference video \(\mathbf Y\), we first construct a restoration loss with pixel, temporal, and frequency constraints:

\[
\begin{aligned}
\mathcal L_{\mathrm{rec}}
={}&\|\hat{\mathbf Y}-\mathbf Y\|_1
+\lambda_t\|\Delta_t\hat{\mathbf Y}-\Delta_t\mathbf Y\|_1\\
&+\lambda_f\left\||\mathcal F_{HW}(\hat{\mathbf Y})|-|\mathcal F_{HW}(\mathbf Y)|\right\|_1,
\end{aligned}
\tag{A.1}
\]

where \(\Delta_t\) denotes inter-frame differences, \(\mathcal F_{HW}\) is the spatial Fourier transform, and \(\lambda_t,\lambda_f\) are the corresponding weights.

An auxiliary loss further constrains loop modulation and expert routing:

\[
\mathcal L_{\mathrm{aux}}
=\lambda_m(\mathcal L_{\mathrm{mask}}+\mathcal R_{\mathrm{overlap}})
+\lambda_b\mathcal R_{\mathrm{load}}+\lambda_c\mathcal L_{\mathrm{cls}},
\tag{A.2}
\]

where the terms constrain adjacent-iteration masks, expert selection overlap, load balancing, and degradation classification, respectively, with weights \(\lambda_m,\lambda_b,\lambda_c\).

The first stage freezes the depth predictor and optimizes the restoration and auxiliary terms in Eq.~\eqref{eq:total-loss} for 100 epochs at four iterations. The superscript \((4)\) denotes losses computed from four-iteration outputs and intermediate states, with cross-iteration momentum updated according to Eq.~\eqref{eq:momentum-update}.

\Needspace{10\baselineskip}
The second stage freezes the restoration network and condition extractor. Candidate depths are evaluated using the quality-cost criterion

\[
\mathcal J(K)=\mathcal L_{\mathrm{rec}}^{(K)}+\eta_t\widetilde T(K)+\eta_m\widetilde M(K),
\tag{A.3}\label{eq:quality-cost}
\]

where \(\widetilde T\) and \(\widetilde M\) are normalized inference time and GPU memory costs, respectively, weighted by \(\eta_t\) and \(\eta_m\).

Using \(\mathcal J(K)\), we assign each training window a target depth \(K^*\):

\[
K^*=\min\!\left(\underset{K\in\{3,4,5\}}{\arg\min}\,\mathcal J(K)\right),
\tag{A.4}
\]
\nopagebreak[4]
where \(K^*\) minimizes the objective, with ties resolved in favor of the smaller \(K\).
\nopagebreak[3]

To learn this selection, \(K^*\) is converted into ordinal labels \(y_4,y_5\), supervising predicted logits \(a_4,a_5\):

\[
y_j=\mathbb I[K^*\geq j],\qquad
\mathcal L_K=\frac12\sum_{j\in\{4,5\}}\operatorname{BCE}(\sigma(a_j),y_j),
\tag{A.5}
\]

where \(y_j\) indicates whether at least \(j\) iterations are required, and \(\operatorname{BCE}\) denotes binary cross-entropy. Only the depth predictor is updated for 100 epochs in this stage:

\[
\mathcal L_{\mathrm{stage2}}=\mathcal L_K.
\tag{A.6}
\]

This training conditions depth selection on input degradation. At inference, the iteration count is predicted directly without reference videos or reevaluation of the quality-cost objective.
\endgroup

\subsection{Inference-Time Depth Selection}\label{app:inference}
The restoration network and condition extractor remain fixed after the first stage. During predictor training, each window is assigned the lowest-cost depth among \(3,4,5\) according to Eq.~\eqref{eq:quality-cost}; ties favor the smaller depth. During inference, two predicted threshold probabilities determine whether the fourth and fifth iterations are executed, as in the main text. Reference frames, quality-cost evaluation, and per-depth reruns are not required at inference.

\section{Experimental Protocol and Reproducibility}\label{app:protocol}

\subsection{Data and Preprocessing}\label{app:data}
Experiments use the UHV-4K-AIO official split of 80 training and 20 test videos across dehazing, deraining, denoising, and low-light enhancement. A single model is used across the four tasks. Training samples aligned \(256\times256\) crops; testing processes five-frame windows and evaluates their center frames at \(3840\times2160\) resolution. All reported task means give equal weight to the four degradation categories.

\subsection{Training Details}\label{app:baselines}
Both training stages run for 100 epochs. Training uses AdamW~\citep{loshchilov2019decoupled}, FP16 mixed precision~\citep{micikevicius2018mixed}, a batch size of 2, gradient clipping, and cosine learning-rate decay with warmup on one NVIDIA V100 (32GB). The restoration network is optimized with four iterations in the first stage; the condition extractor and restoration network are frozen when training the depth predictor in the second stage.

\subsection{Metrics and Temporal Evaluation}\label{app:metrics}
Spatial reconstruction is reported with PSNR and SSIM; perceptual quality is reported with LPIPS, MUSIQ, and MANIQA. Temporal consistency is evaluated using dataset-provided reference optical flow for Warping Error and LPIPS features for tLPIPS. Lower Warping Error and tLPIPS indicate better temporal consistency. The complete temporal comparison appears in Appendix~\ref{app:dynamic-temporal}.

\subsection{Time-Varying Degradations and Temporal Consistency}\label{app:dynamic-temporal}
Table~\visualref{tab:switching} reports restoration quality when the degradation type switches every \(\Delta t\in\{6,3\}\) frames. Table~\visualref{tab:temporal} compares temporal stability averaged across the four tasks; LoopMoEVR uses its predicted iteration depth.

\begin{table}[!htbp]
\centering
\small
\setlength{\tabcolsep}{5pt}
\renewcommand{\arraystretch}{1.22}
\begin{adjustbox}{max width=\linewidth}
\begin{tabularx}{\linewidth}{@{}>{\raggedright\arraybackslash}p{.23\linewidth}>{\raggedright\arraybackslash}p{.14\linewidth}|>{\raggedleft\arraybackslash}X|>{\raggedleft\arraybackslash}X>{\raggedleft\arraybackslash}X|>{\raggedleft\arraybackslash}X>{\raggedleft\arraybackslash}X@{}}
\toprule
\multirow{2}{*}{Method} & \multirow{2}{*}{Venue} & \multicolumn{1}{c|}{Static} & \multicolumn{2}{c|}{$\Delta t=6$} & \multicolumn{2}{c}{$\Delta t=3$} \\
\cmidrule(lr){3-3}\cmidrule(lr){4-5}\cmidrule(l){6-7}
 & & PSNR$\uparrow$ & PSNR$\uparrow$ & SSIM$\uparrow$ & PSNR$\uparrow$ & SSIM$\uparrow$ \\
\midrule
\rowcolor{black!8}\multicolumn{7}{c}{\textit{(a) Image methods}} \\
\midrule
AirNet & CVPR'22 & 27.11 & 26.09 & 0.914 & 25.76 & 0.915 \\
TransWeather & CVPR'22 & 25.87 & 23.41 & 0.834 & 23.27 & 0.832 \\
DRNet & TMM'26 & 18.62 & 23.17 & 0.820 & 23.05 & 0.824 \\
UHD-processer & CVPR'25 & 26.77 & 24.05 & 0.746 & 24.19 & 0.752 \\
R2R & CVPR'26 & 15.75 & 26.02 & 0.829 & 25.59 & 0.828 \\
\midrule
\rowcolor{black!8}\multicolumn{7}{c}{\textit{(b) Video methods}} \\
\midrule
EDVR & CVPRW'19 & 21.96 & 20.15 & 0.674 & 24.95 & 0.727 \\
BasicVSR++ & CVPR'22 & 26.25 & 23.87 & 0.864 & 25.81 & 0.885 \\
Shift-Net & CVPR'23 & 26.87 & 25.62 & 0.799 & 27.14 & 0.821 \\
RVRT & NeurIPS'22 & 20.13 & 19.49 & 0.790 & 21.86 & 0.809 \\
ViWS-Net & ICCV'23 & 20.71 & 21.52 & 0.812 & 22.28 & 0.832 \\
AverNet & NeurIPS'24 & 26.92 & 22.36 & 0.714 & 24.94 & 0.756 \\
MoCRA & arxiv'2026 & 27.23 & 26.89 & 0.898 & 27.16 & 0.901 \\
\midrule
\rowcolor{black!5}
\textbf{LoopMoEVR (Ours)} & -- & \textbf{27.87} & \textbf{27.68} & \textbf{0.916} & \textbf{27.85} & \textbf{0.915} \\
\bottomrule
\end{tabularx}
\end{adjustbox}
\normalsize
\caption{Average restoration performance under time-varying degradations.}
\label{tab:switching}
\end{table}

\begin{table}[!htbp]
\centering
\footnotesize
\setlength{\tabcolsep}{5pt}
\renewcommand{\arraystretch}{1.05}
\begin{tabular}{@{}llrr@{}}
\toprule
Method & Venue & Warping Error$\downarrow$ & tLPIPS$\downarrow$ \\
\midrule
AirNet & CVPR'22 & 0.0305 & 0.1256 \\
TransWeather & CVPR'22 & 0.0329 & 0.1420 \\
UHD-processer & CVPR'25 & 0.0327 & 0.1551 \\
\midrule
AverNet & NeurIPS'24 & \textbf{0.0296} & 0.1238 \\
MoCRA & arxiv'2026 & 0.0300 & 0.1375 \\
\rowcolor{black!5}
\textbf{LoopMoEVR (Ours)} & -- & 0.0297 & \textbf{0.1182} \\
\bottomrule
\end{tabular}
\caption{Temporal stability averaged across four tasks. Warping Error is computed using the same reference optical flow for all methods.}
\label{tab:temporal}
\end{table}

\FloatBarrier

\subsection{Non-UHD Resolution Results}\label{app:non-uhd-results}
Table~\visualref{tab:non-uhd-results} reports PSNR and SSIM on MOT17 and DAVIS-test under two evolving-degradation settings~\citep{li2026allinonevideorestorationsmoothly}. Setting 1 varies degradation severity within each video, while Setting 2 also changes the degradation type. LoopMoEVR and the comparison methods were trained and evaluated under the same respective settings. Average denotes the arithmetic mean of Settings 1 and 2 for each dataset.

\begin{table}[!htbp]
\centering
\caption{Restoration results on MOT17 and DAVIS-test under two evolving-degradation settings.}
\label{tab:non-uhd-results}
\begingroup
\setlength{\tabcolsep}{2.2pt}
\renewcommand{\arraystretch}{1.13}
\begin{adjustbox}{max width=\linewidth}
\begin{tabular}{@{}l|rr|rr|rr|rr|rr|rr@{}}
\toprule
\multirow{3}{*}{Method} & \multicolumn{6}{c|}{MOT17} & \multicolumn{6}{c}{DAVIS-test} \\
\cmidrule(lr){2-7}\cmidrule(l){8-13}
 & \multicolumn{2}{c|}{Setting 1} & \multicolumn{2}{c|}{Setting 2} & \multicolumn{2}{c|}{Average} & \multicolumn{2}{c|}{Setting 1} & \multicolumn{2}{c|}{Setting 2} & \multicolumn{2}{c}{Average} \\
\cmidrule(lr){2-3}\cmidrule(lr){4-5}\cmidrule(lr){6-7}\cmidrule(lr){8-9}\cmidrule(lr){10-11}\cmidrule(l){12-13}
 & PSNR & SSIM & PSNR & SSIM & PSNR & SSIM & PSNR & SSIM & PSNR & SSIM & PSNR & SSIM \\
\midrule
TransWeather & 18.73 & 0.5506 & 19.94 & 0.6145 & 19.34 & 0.5826 & 20.44 & 0.5672 & 20.64 & 0.5839 & 20.54 & 0.5756 \\
AirNet & 23.62 & 0.8900 & 26.56 & 0.9412 & 25.09 & 0.9156 & 24.65 & 0.9034 & 19.93 & 0.8774 & 22.29 & 0.8904 \\
BasicVSR++ & 36.26 & 0.9793 & 32.43 & 0.9787 & 34.35 & 0.9790 & 32.33 & 0.9739 & 25.51 & 0.9527 & 28.92 & 0.9633 \\
RVRT & 25.63 & 0.8851 & 23.36 & 0.9040 & 24.50 & 0.8946 & 27.71 & 0.9225 & 19.39 & 0.8313 & 23.55 & 0.8769 \\
AverNet & 28.98 & 0.8894 & 27.42 & 0.8582 & 28.20 & 0.8738 & 25.46 & 0.8023 & 25.01 & 0.7952 & 25.24 & 0.7988 \\
\rowcolor{black!5}
\textbf{LoopMoEVR (Ours)} & 31.95 & 0.9184 & 30.89 & 0.9070 & 31.42 & 0.9127 & 27.61 & 0.8921 & 26.75 & 0.8831 & 27.18 & 0.8876 \\
\bottomrule
\end{tabular}
\end{adjustbox}
\endgroup
\end{table}

\FloatBarrier

\subsection{Anonymous User Study}\label{app:user-study}
\begin{figure}[!htbp]
\centering
\begin{minipage}{.92\linewidth}
\vspace{0pt}
\centering
\includegraphics[width=\linewidth]{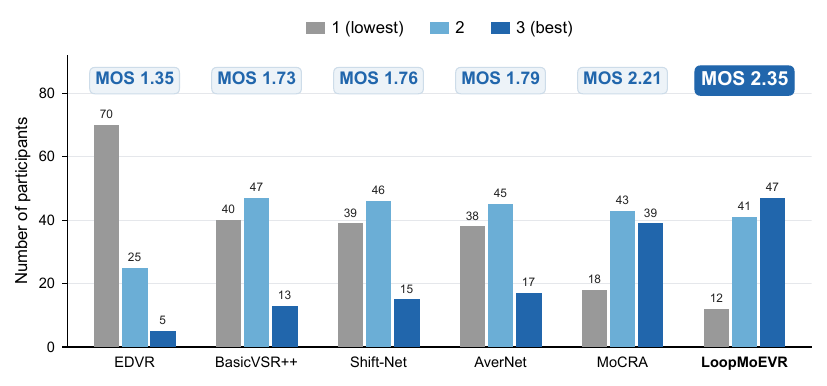}
\captionof{figure}{User-study rating distribution (100 participants; 3 = best). Mean opinion scores (MOS) are shown above each method.}
\label{fig:user-study}
\end{minipage}
\end{figure}

One hundred participants anonymously rated six methods on a common three-point visual-quality scale (3 = best), considering local details, color and brightness, and temporal consistency. LoopMoEVR received the most top ratings (47/100) and the highest mean opinion score (MOS 2.35), compared with MoCRA (39/100; MOS 2.21). Figure~\visualref{fig:user-study} shows the rating distribution and MOS for each method.

\section{Extended Ablations}\label{app:extended}

\subsection{Core Components}\label{app:core-ablation}
Table~\visualref{tab:ablation-summary}(a) in the main paper summarizes the core-component ablations. Except for depth allocation, variants are independently trained for four iterations with the same data split and training budget. Video-conditioning ablations remove only the corresponding pooled-state and momentum inputs while retaining local-feature and iteration conditioning. A qualitative comparison appears in Figure~\visualref{fig:ablation-core-visual}.

\subsection{Loop-Conditioning Formulations}\label{app:conditioning}
Table~\visualref{tab:ablation-summary}(c) in the main paper compares the base iteration encoding, additive degradation conditioning, a concatenation MLP, and the low-rank interaction. The three degradation-conditioned alternatives have matched conditioning-generator parameter budgets and residual-output initialization. The corresponding visualization appears in Figure~\visualref{fig:ablation-components-conditioning}.

\subsection{Expert-Branch Architecture Ablation}\label{app:expert-branch-ablation}
To evaluate the expert-branch architecture, we fix the iteration count at \(K=4\) and compare two dense replacement branches, a branch retaining only routed experts, and the full model under the same training and evaluation settings. Table~\visualref{tab:expert-branch-ablation} shows that the routed-only variant outperforms both dense alternatives. Adding shared experts further yields the best result among the evaluated configurations, improving PSNR by 1.03 dB over the routed-only variant. Within the tested architectures, the full expert branch achieves stronger overall restoration performance, supporting the effectiveness of jointly using routed and shared experts for unified video restoration.

\begin{table}[!htbp]
\centering
\begin{minipage}{.88\linewidth}
\setlength{\tabcolsep}{4pt}
\renewcommand{\arraystretch}{1.2}
\fontsize{8}{9.2}\selectfont
\begin{tabularx}{\linewidth}{@{}>{\raggedright\arraybackslash}Xrrr@{}}
\toprule
Expert branch & PSNR$\uparrow$ & LPIPS$\downarrow$ & tLPIPS$\downarrow$ \\
\midrule
Compact Dense & 25.83 & 0.162 & 0.1484 \\
Expanded Dense & 26.67 & 0.153 & 0.1392 \\
Routed experts only & 26.95 & 0.137 & 0.1289 \\
\midrule
\rowcolor{black!5}
\textbf{Full model (Ours)} & \textbf{27.98} & \textbf{0.122} & \textbf{0.1144} \\
\bottomrule
\end{tabularx}
\par\smallskip
\setlength{\abovecaptionskip}{0pt}
\fontsize{8}{9.2}\selectfont
\captionof{table}{Expert-branch architecture ablation at \(K=4\).}
\label{tab:expert-branch-ablation}
\end{minipage}
\end{table}

\subsection{Loop-Depth Allocation}\label{app:depth-ablation}
Table~\visualref{tab:ablation-summary}(b) in the main paper compares fixed, random, and input-adaptive depth policies using the same four-iteration checkpoint. The random and dynamic policies have the same within-task depth distribution. Relative to \(K=3\), \(K=2\) saves 87 ms but loses 0.89 dB PSNR and increases tLPIPS by 0.0300. Relative to \(K=5\), \(K=6\) costs 98 ms for a 0.03 dB PSNR gain and slightly higher tLPIPS. Considering the restoration quality and inference time of the fixed-depth variants, we restrict the adaptive depth candidates to \(\{3,4,5\}\).

\subsection{Loss-Function Components}\label{app:loss-ablation}
Table~\visualref{tab:ablation-summary}(d) in the main paper reports independently trained four-iteration variants with individual loss terms omitted while the remaining weights are held fixed. Removing temporal supervision primarily weakens tLPIPS; removing frequency or auxiliary terms reduces PSNR and perceptual quality. The visual comparison appears in Figure~\visualref{fig:ablation-loss} of the main paper.

\subsection{Ablation of Looped Temporal Modeling}\label{app:temporal-modeling-ablation}
To assess the effectiveness of our looped temporal modeling, we fix the iteration count at \(K=4\) and modify how state conditions are constructed and used for modulation under the same training settings. We also examine the joint removal of the momentum state and temporal loss \(\mathcal L_t\). Table~\visualref{tab:temporal-modeling-ablation} shows that the full model achieves the best PSNR, Warping Error, and tLPIPS among the configurations considered. Across the four state-path variants, PSNR decreases by 0.78--0.89 dB and tLPIPS increases by 0.0110--0.0160, supporting the contribution of our looped state modeling to reconstruction accuracy and temporal consistency. Removing both the momentum state and \(\mathcal L_t\) further reduces PSNR to 26.85 dB and increases tLPIPS to 0.1744, indicating that these two components jointly support temporally stable video restoration.

\begin{table}[!htbp]
\centering
\begin{minipage}{.88\linewidth}
\setlength{\tabcolsep}{4pt}
\renewcommand{\arraystretch}{1.2}
\fontsize{8}{9.2}\selectfont
\begin{tabularx}{\linewidth}{@{}>{\raggedright\arraybackslash}Xrrr@{}}
\toprule
Configuration & PSNR$\uparrow$ & \shortstack{Warping\\Error$\downarrow$} & tLPIPS$\downarrow$ \\
\midrule
w/o current pooled condition & 27.14 & 0.0312 & 0.1274 \\
w/o momentum history & 27.09 & 0.0317 & 0.1304 \\
Modulation with pre-update state & 27.20 & 0.0308 & 0.1254 \\
w/o loop mask & 27.17 & 0.0310 & 0.1264 \\
w/o momentum state and \(\mathcal L_t\) & 26.85 & 0.0375 & 0.1744 \\
\midrule
\rowcolor{black!5}
\textbf{Full model (Ours)} & \textbf{27.98} & \textbf{0.0293} & \textbf{0.1144} \\
\bottomrule
\end{tabularx}
\par\smallskip
\setlength{\abovecaptionskip}{0pt}
\fontsize{8}{9.2}\selectfont
\captionof{table}{Ablation of looped temporal modeling at \(K=4\).}
\label{tab:temporal-modeling-ablation}
\end{minipage}
\end{table}
\FloatBarrier

\subsection{Cross-Iteration Parameter Sharing Ablation}\label{app:parameter-sharing}
To evaluate cross-iteration parameter sharing, we compare shared and independent parameter configurations at matched network width and a fixed depth of four iterations. As shown in Table~\visualref{tab:parameter-sharing}, sharing reduces the parameter count from 3.116M to 0.884M, a reduction of 71.6\%, with a PSNR difference of only 0.08 dB and a lower tLPIPS. Through cross-iteration parameter reuse, LoopMoEVR maintains restoration accuracy close to the independent configuration while improving temporal consistency, demonstrating the parameter efficiency of looped computation for unified UHD video restoration.

\begin{table}[!htbp]
\centering
\begin{minipage}{.88\linewidth}
\setlength{\tabcolsep}{4pt}
\renewcommand{\arraystretch}{1.2}
\fontsize{8}{9.2}\selectfont
\begin{tabularx}{\linewidth}{@{}>{\raggedright\arraybackslash}Xrrrr@{}}
\toprule
Parameter configuration & Params/M$\downarrow$ & $\Delta$PSNR/dB$\uparrow$ & tLPIPS$\downarrow$ & Time/ms$\downarrow$ \\
\midrule
Unshared across iterations & 3.116 & \textbf{+0.08} & 0.1160 & \textbf{515} \\
\rowcolor{black!5}
\textbf{Shared across iterations (Ours)} & \textbf{0.884} & 0.00 & \textbf{0.1144} & 560 \\
\bottomrule
\end{tabularx}
\par\smallskip
\setlength{\abovecaptionskip}{0pt}
\fontsize{8}{9.2}\selectfont
\captionof{table}{Cross-iteration parameter sharing ablation. All configurations use four iterations; $\Delta$PSNR is relative to the shared configuration, and time is end-to-end latency per five-frame 4K window.}
\label{tab:parameter-sharing}
\end{minipage}
\end{table}
\FloatBarrier

\section{Qualitative Evidence and Mechanism Analysis}\label{app:qualitative}

\subsection{Visual Comparisons Retained in the Main Paper}\label{app:main-visuals}
The main paper retains the full six-method and identical-crop comparison across four restoration tasks (Figure~\visualref{fig:qualitative}), the smartphone demonstration (Figure~\visualref{fig:intro-mobile}), and the iteration-by-iteration dehazing sequence (Figure~\visualref{fig:discussion-iterations}). These displays document complementary aspects of the result: comparative local detail, a deployment example, and progressive correction within a shared-parameter backbone.

\subsection{Qualitative Ablation Comparisons}\label{app:ablation-vis}
\begin{figure}[!htbp]
\centering
\noindent\ablationtile{.242\linewidth}{a}{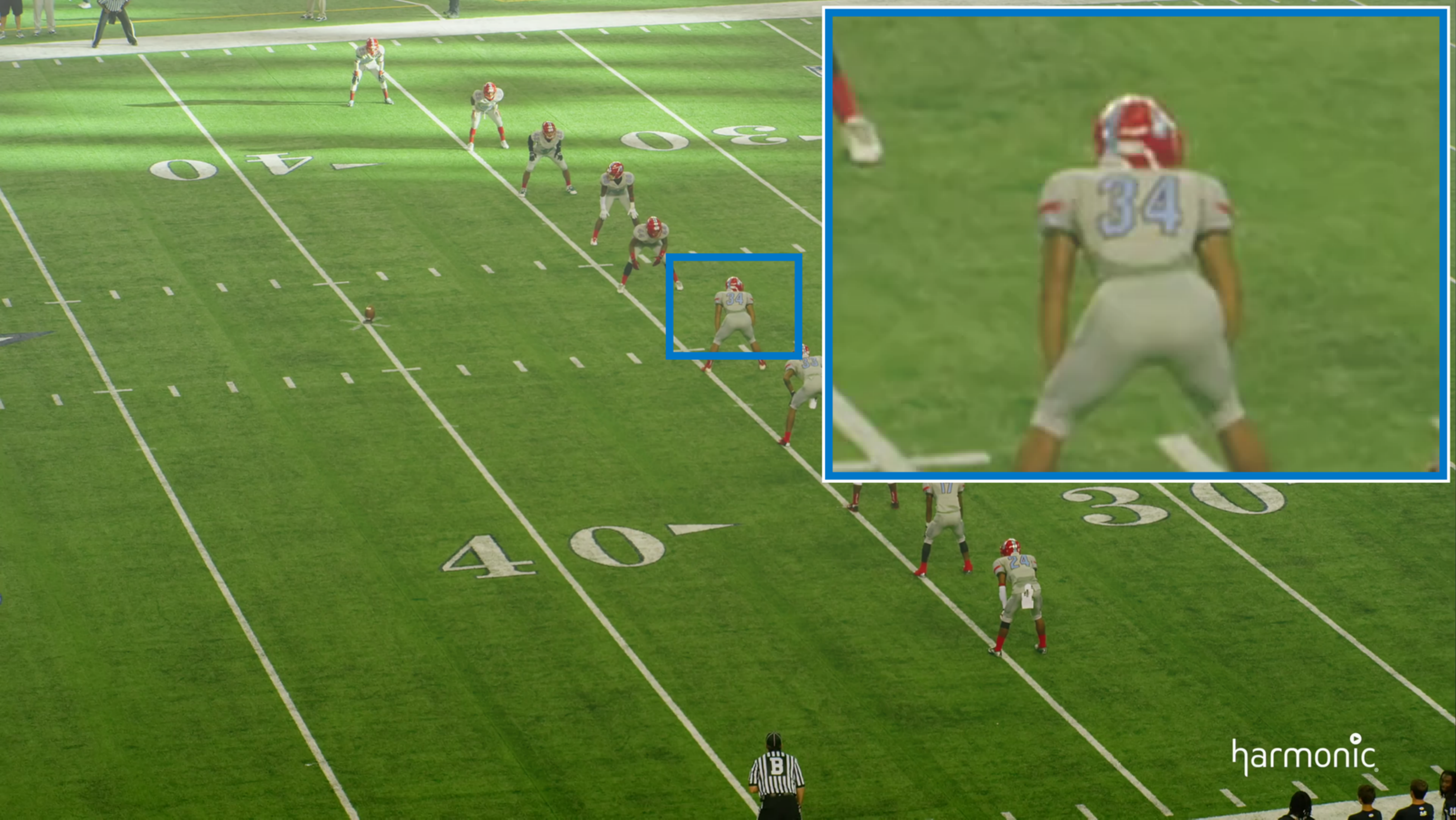}{Input}\hfill%
\ablationtile{.242\linewidth}{b}{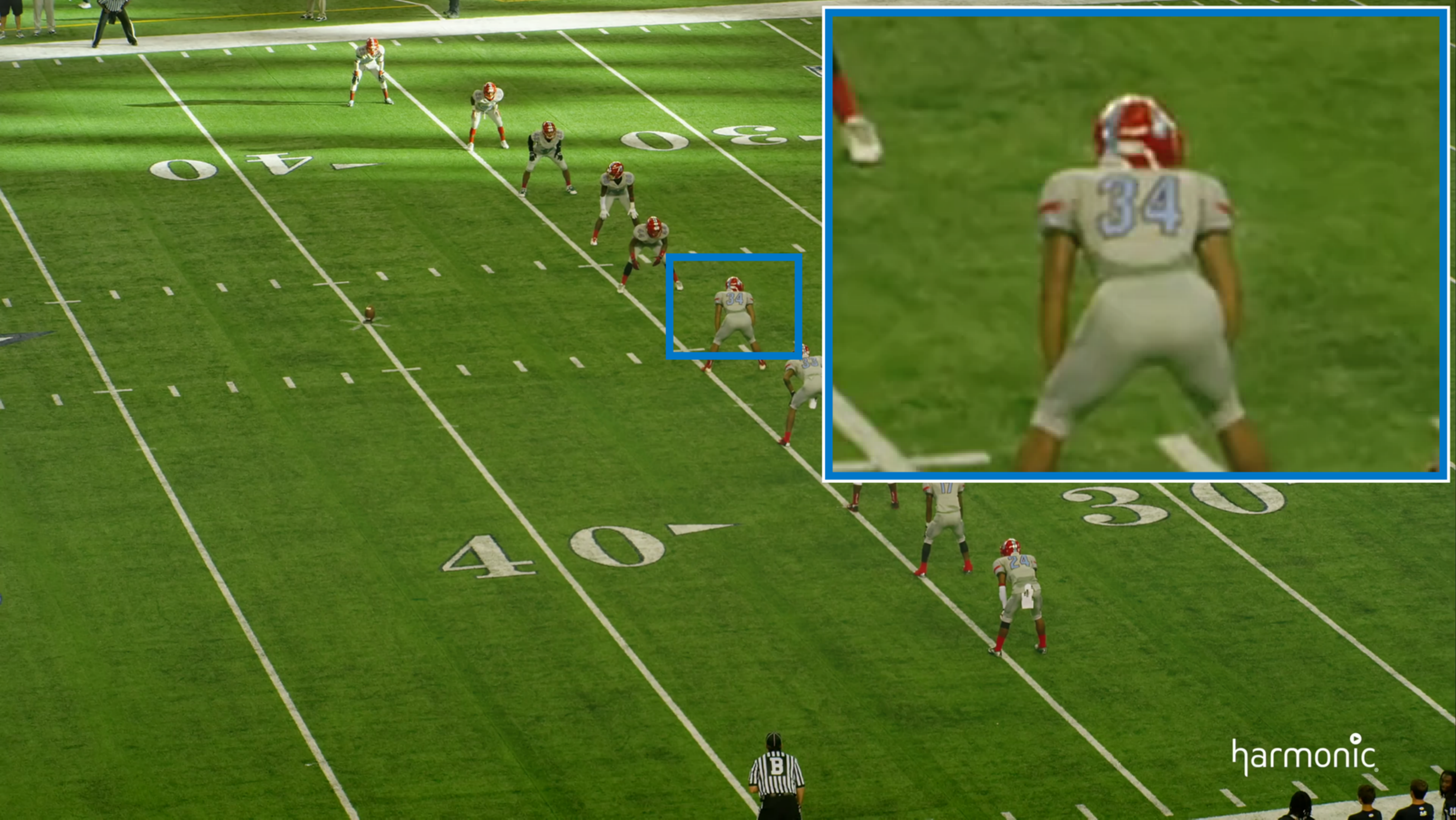}{LoopMoE-style\newline video baseline}\hfill%
\ablationtile{.242\linewidth}{c}{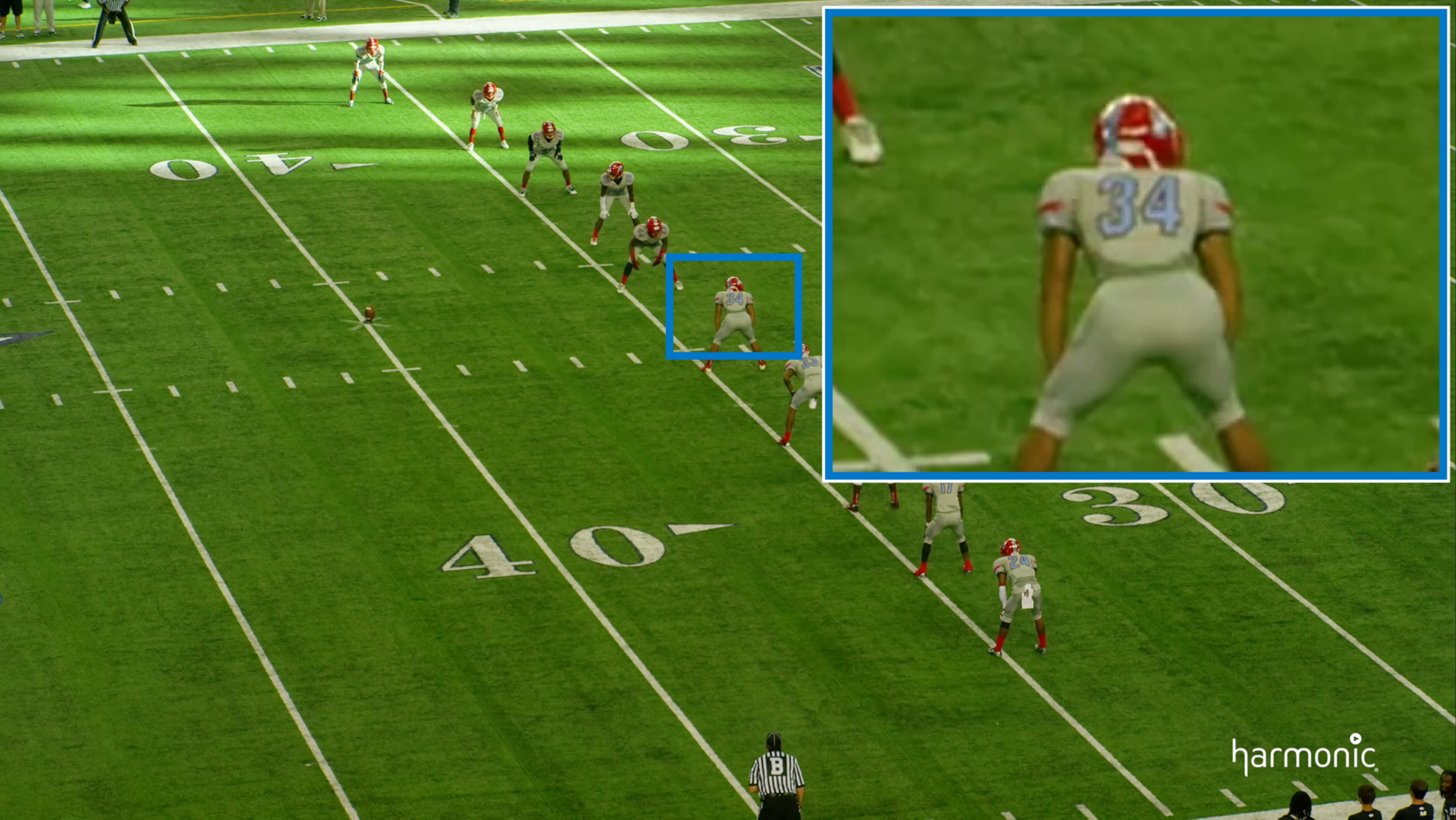}{w/o degradation\newline residual}\hfill%
\ablationtile{.242\linewidth}{d}{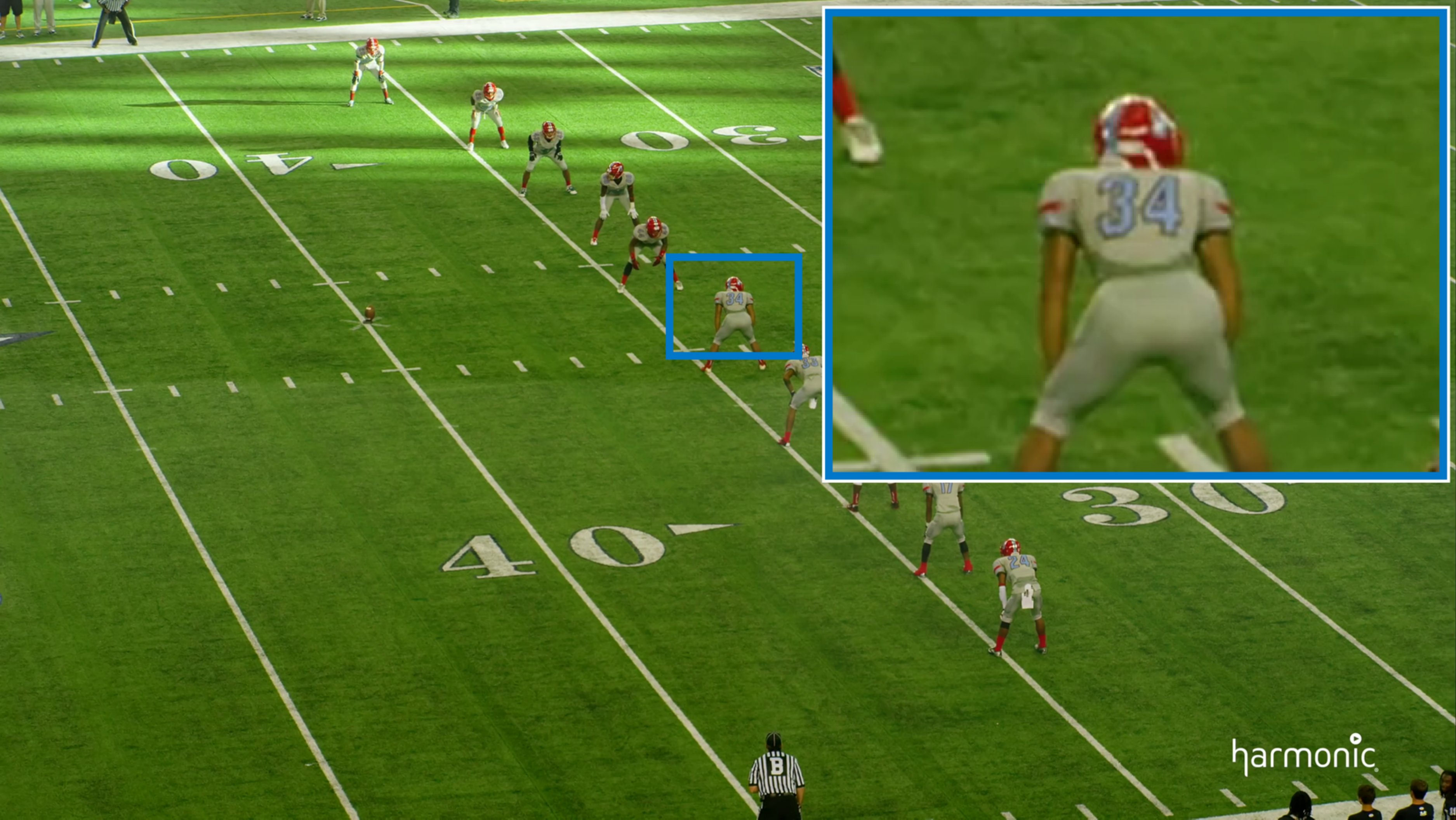}{w/o dual-branch\newline video conditioning}\par\vspace{3pt}
\noindent\ablationtile{.242\linewidth}{e}{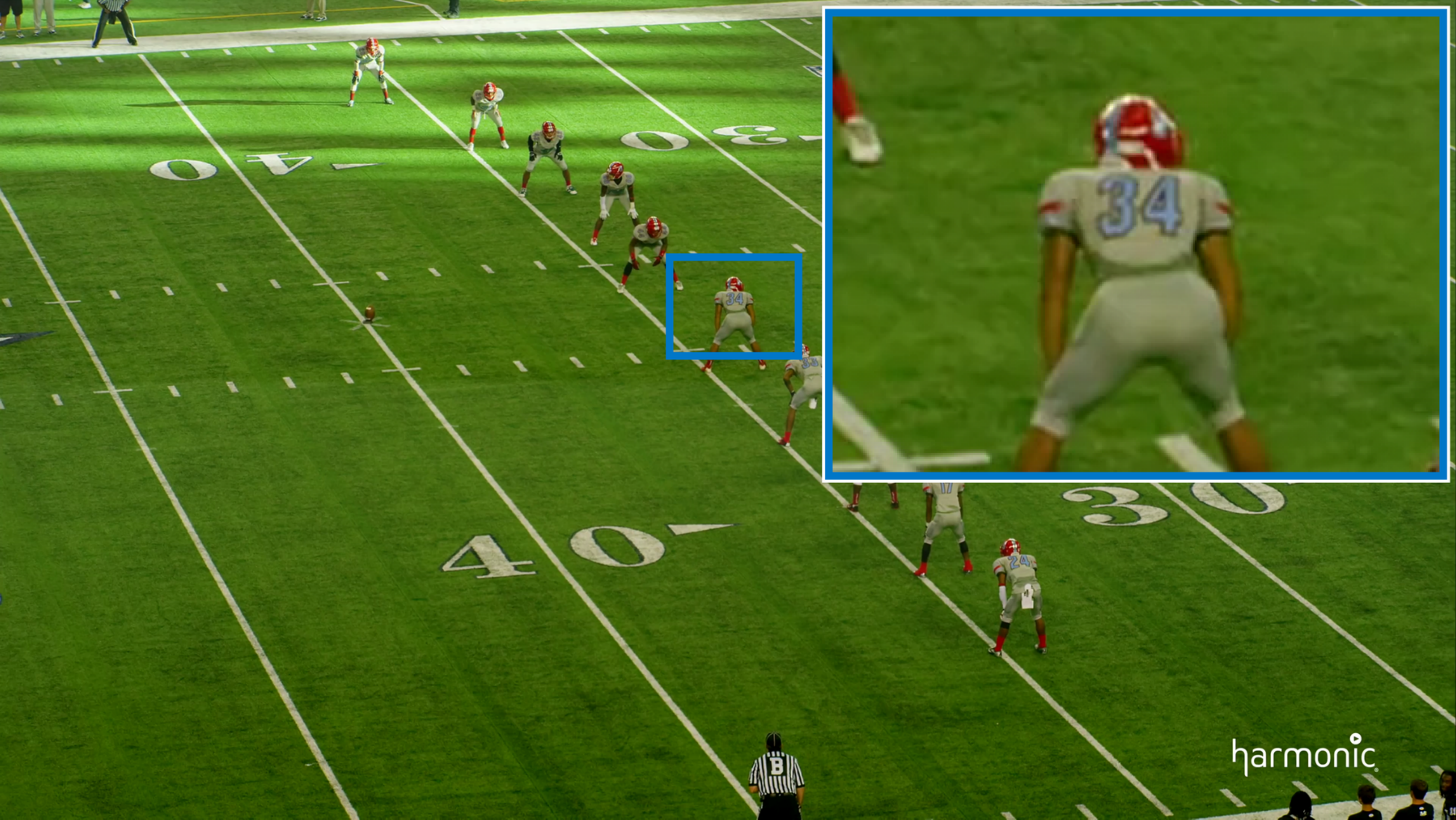}{w/o expert video\newline conditioning}\hfill%
\ablationtile{.242\linewidth}{f}{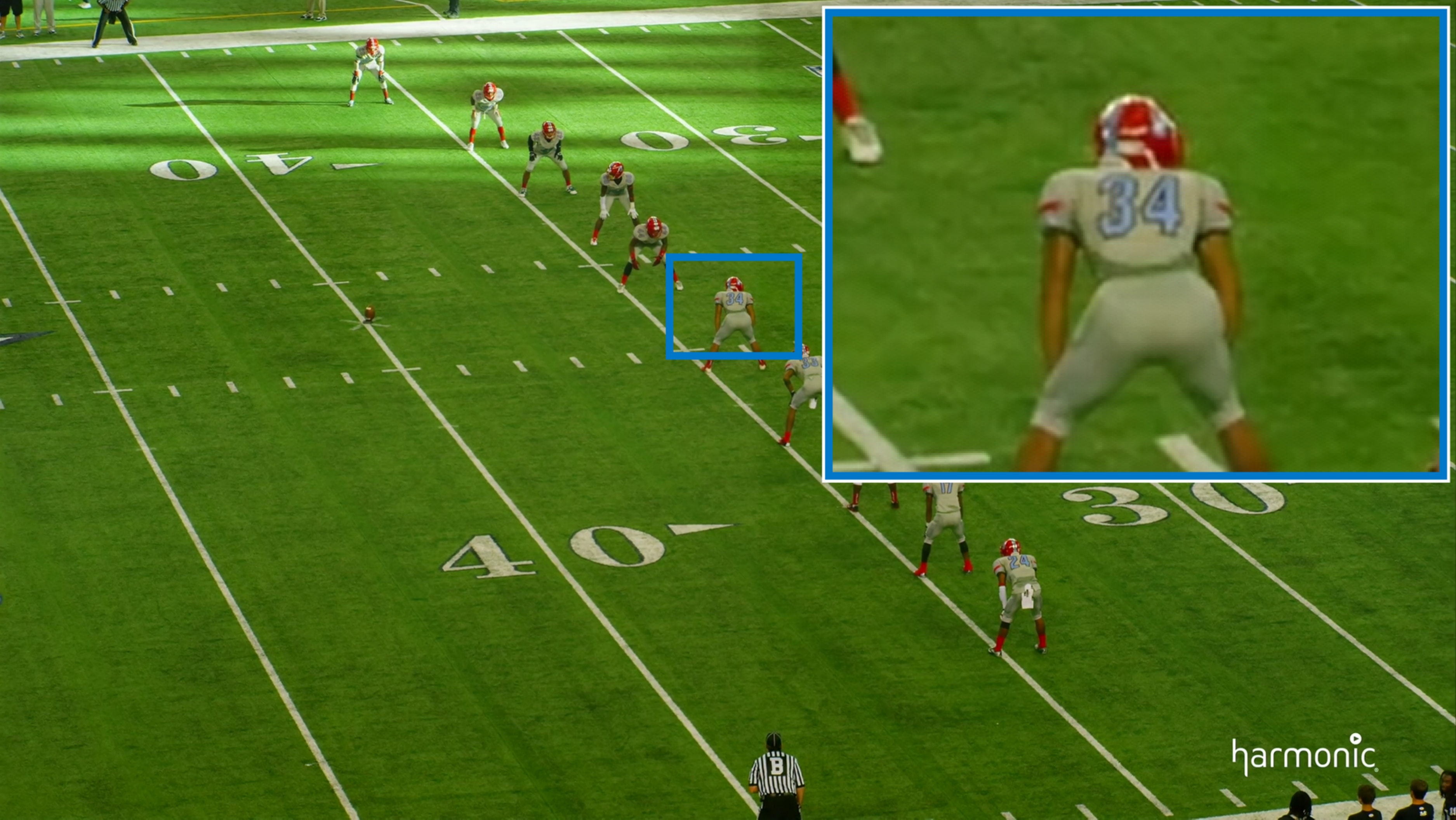}{w/o momentum\newline state}\hfill%
\ablationtile{.242\linewidth}{g}{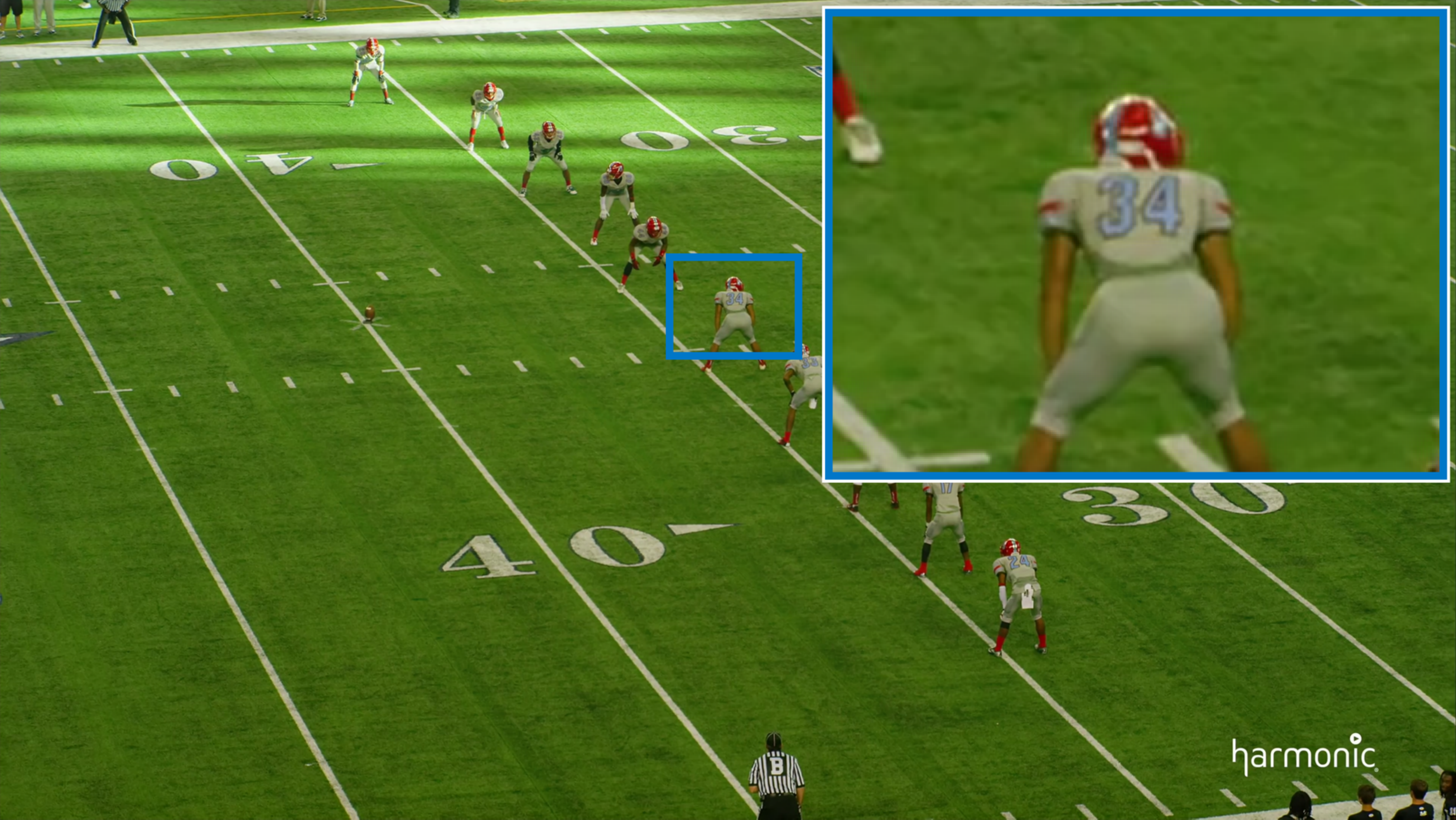}{\textbf{Ours}}\hfill%
\ablationtile{.242\linewidth}{h}{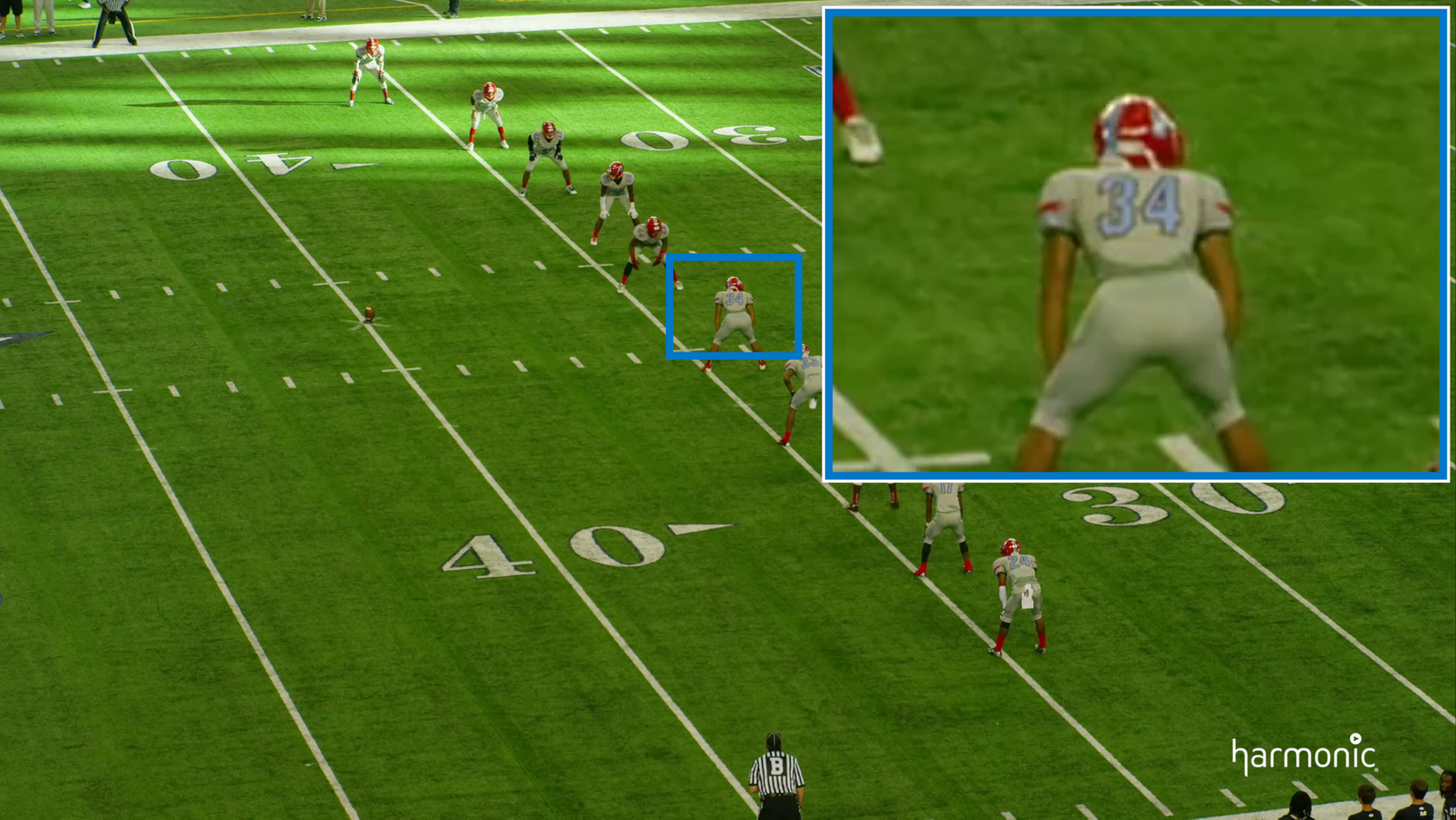}{GT}
\caption{Qualitative comparison of core-component ablations.}
\label{fig:ablation-core-visual}
\end{figure}

Figure~\visualref{fig:ablation-core-visual} compares the core-component variants, and Figure~\visualref{fig:ablation-components-conditioning} compares loop-conditioning formulations. The loss-function visualization appears in the main paper (Figure~\visualref{fig:ablation-loss}).

\begin{center}
\noindent\ablationtile{.16\linewidth}{a}{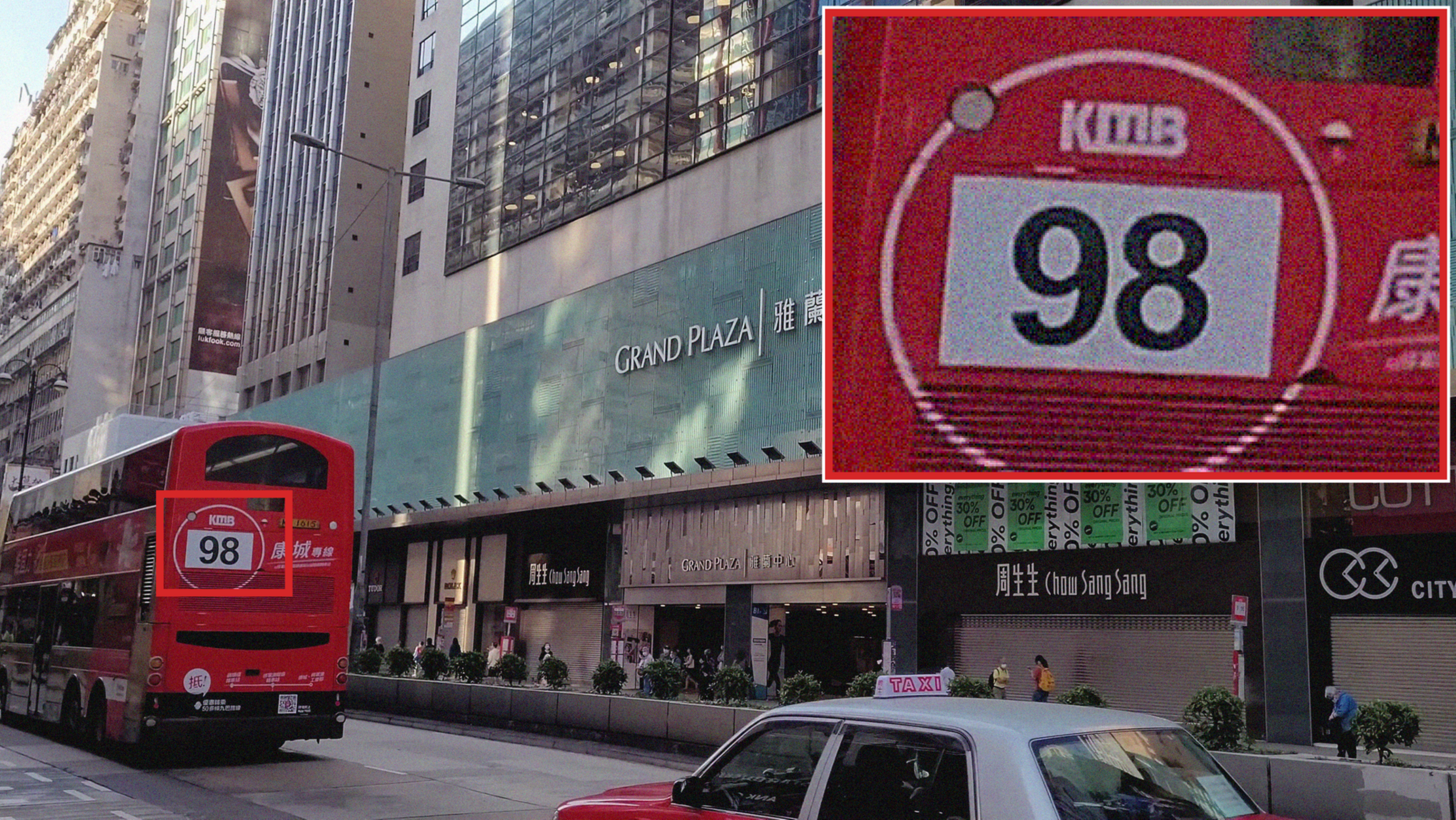}{Input}\hfill%
\ablationtile{.16\linewidth}{b}{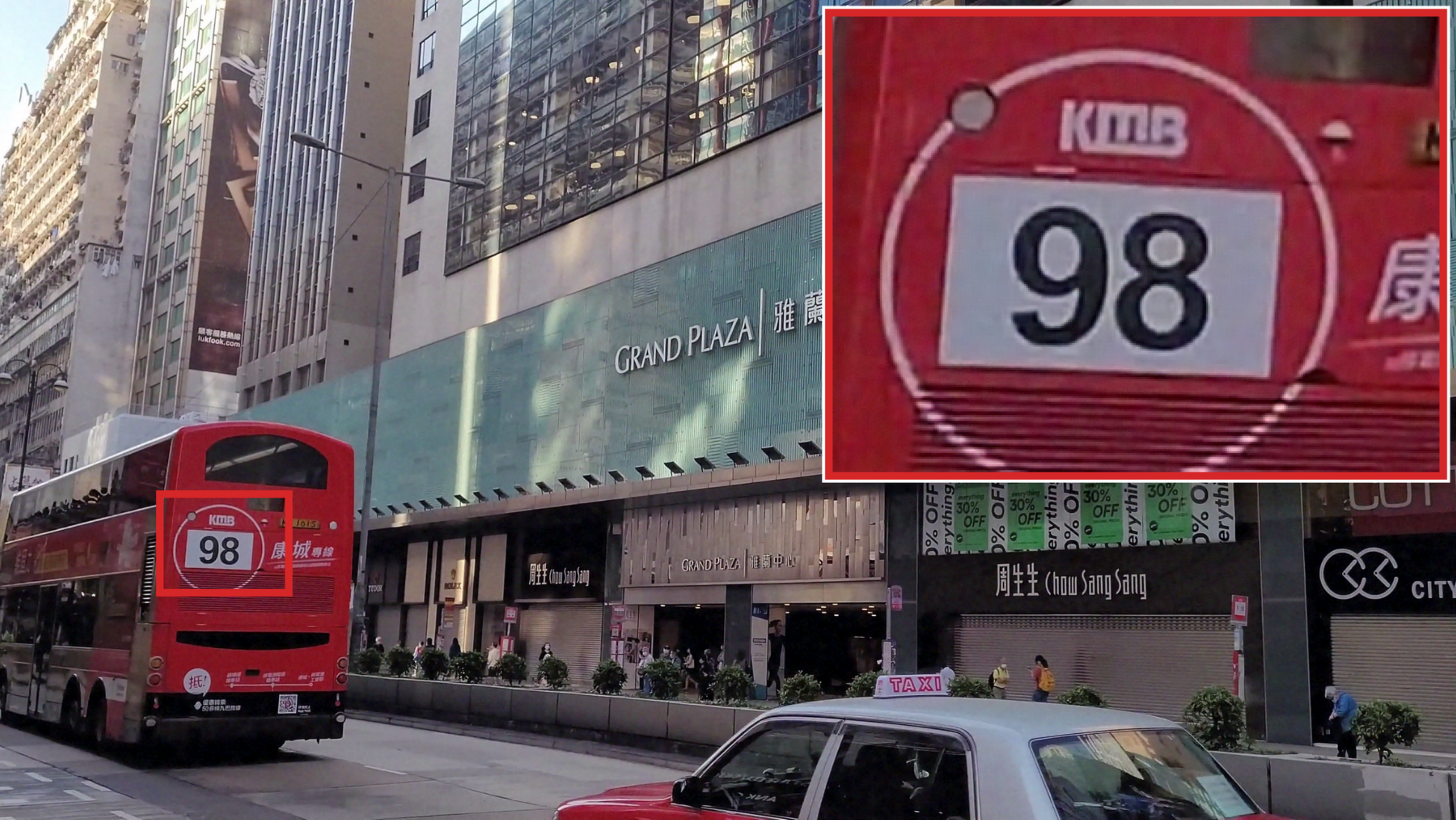}{Base iteration\newline encoding}\hfill%
\ablationtile{.16\linewidth}{c}{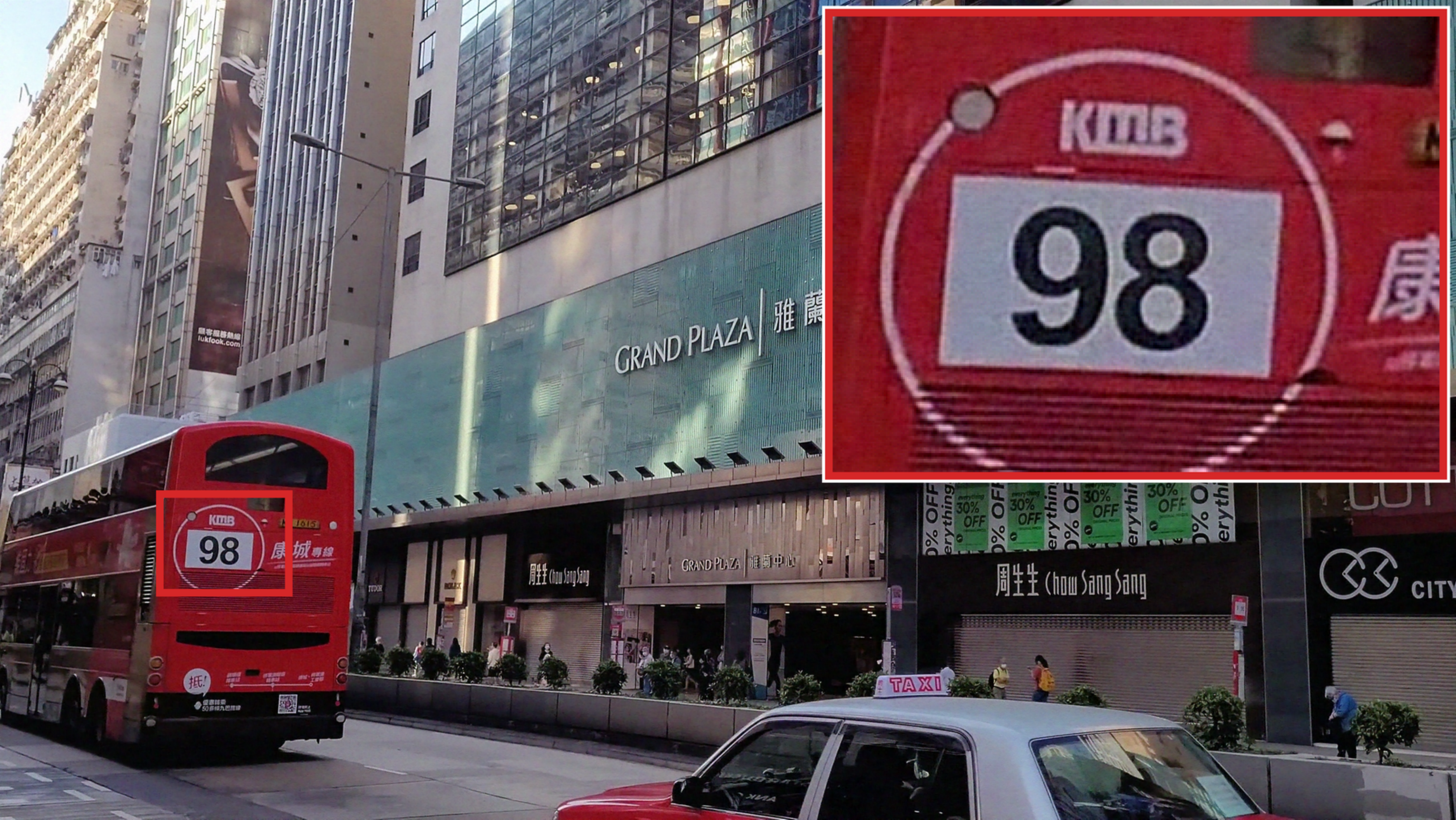}{Additive degradation\newline conditioning}\hfill%
\ablationtile{.16\linewidth}{d}{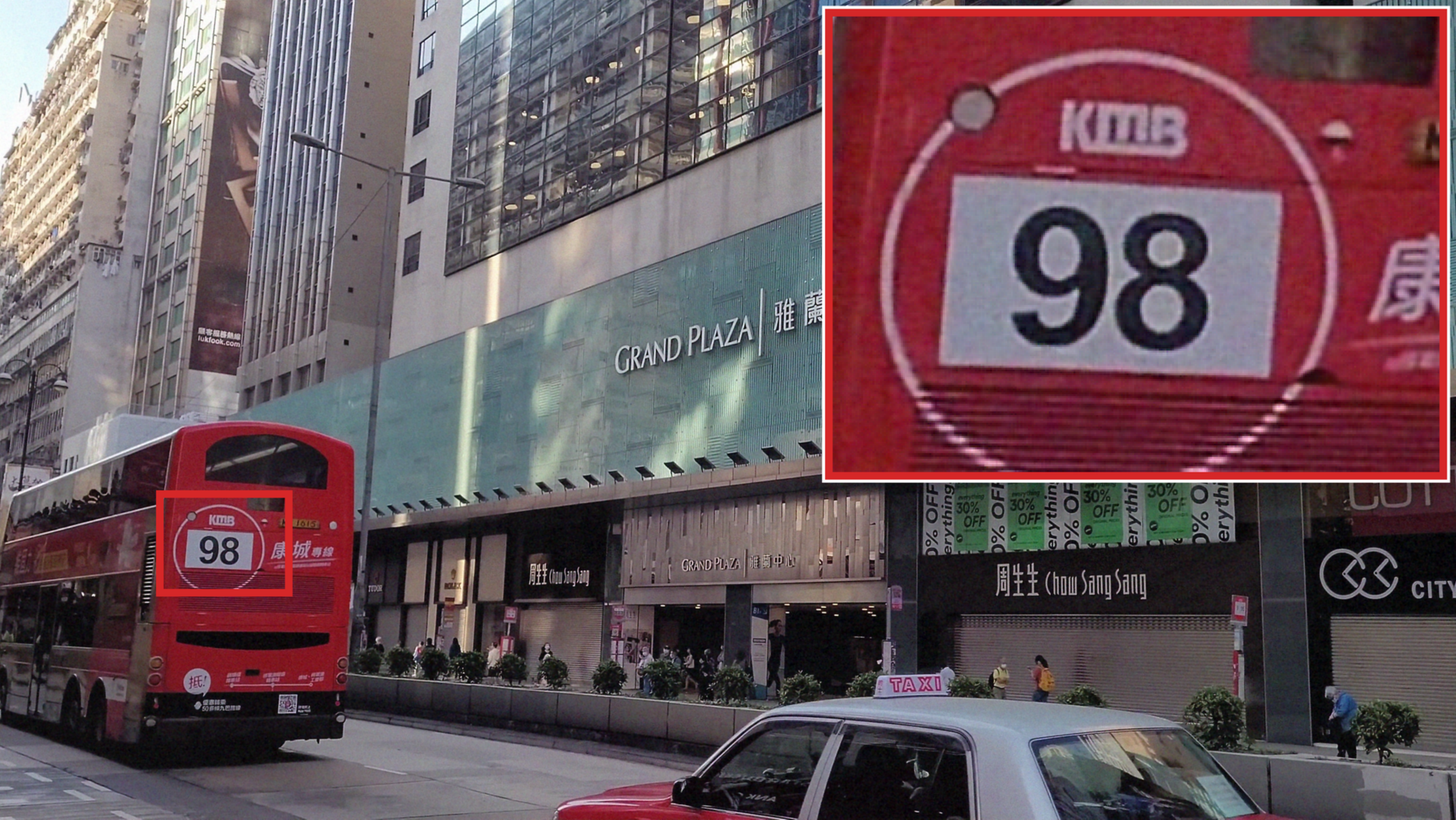}{Concatenation\newline MLP}\hfill%
\ablationtile{.16\linewidth}{e}{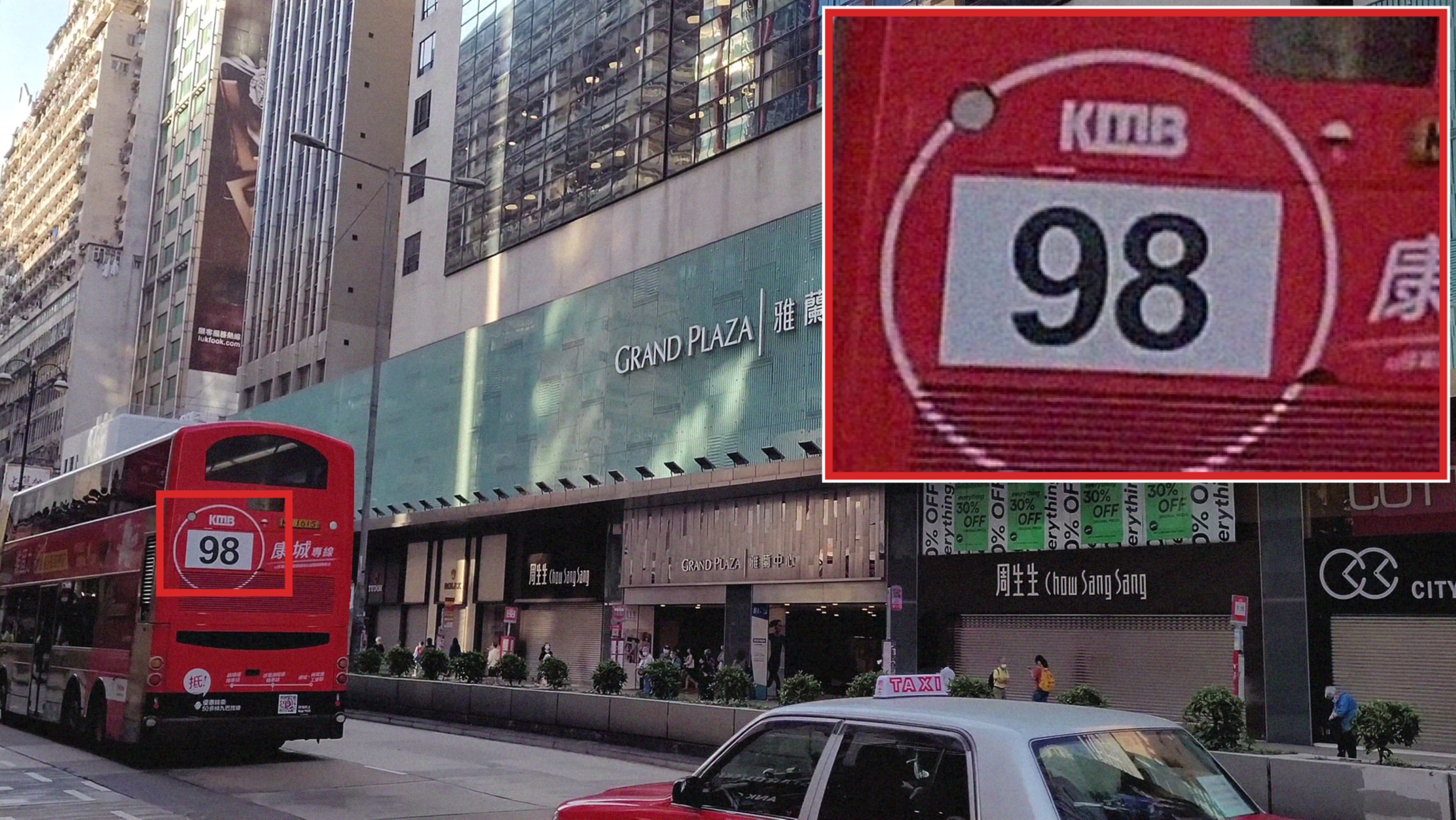}{\textbf{Ours}}\hfill%
\ablationtile{.16\linewidth}{f}{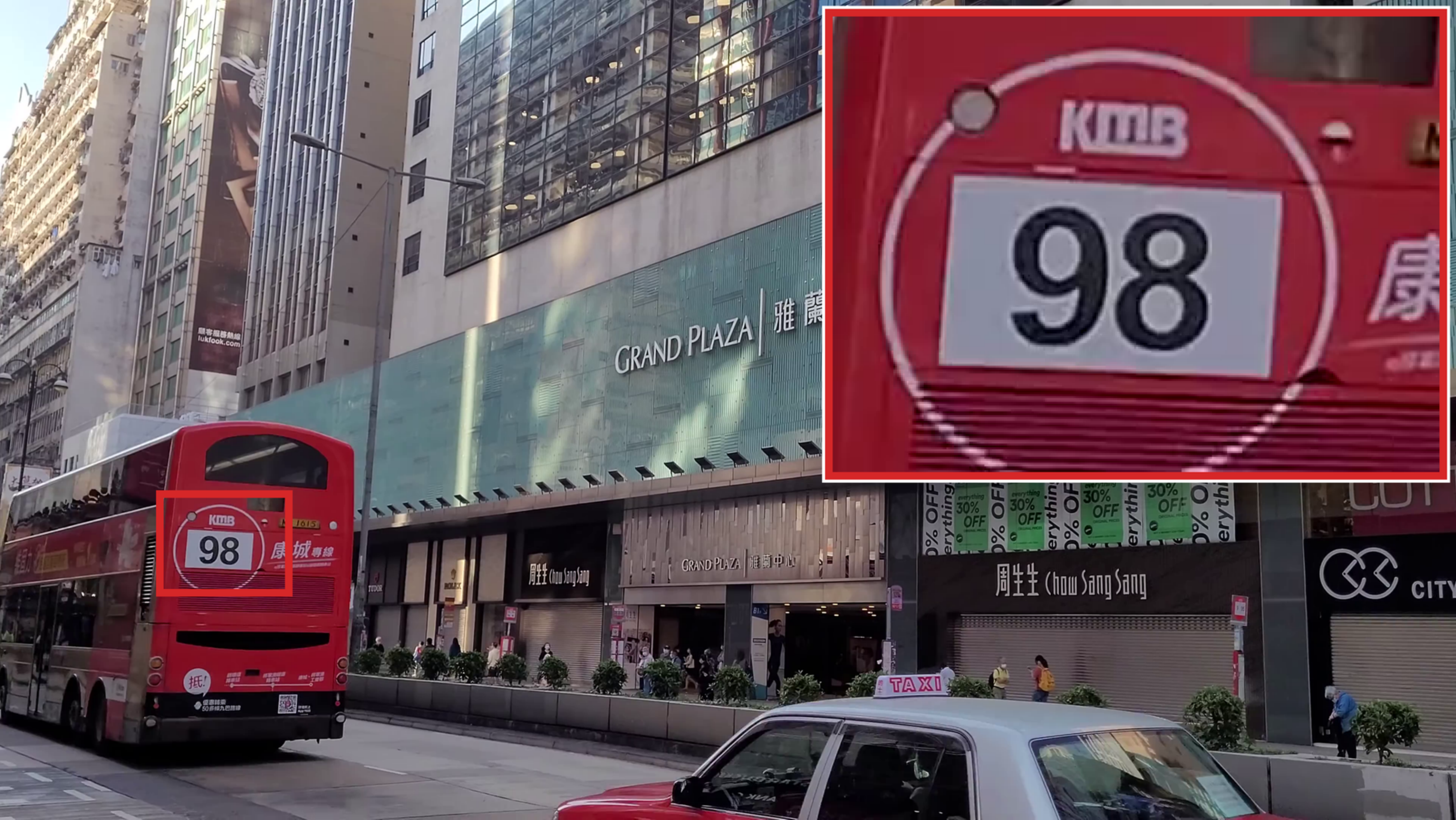}{GT}
\captionof{figure}{Qualitative comparison of loop-conditioning ablations.}
\label{fig:ablation-components-conditioning}
\end{center}

\FloatBarrier
\subsection{Loop-Condition Order}\label{app:condition-order}
\begin{figure}[!htbp]
\centering
\includegraphics[width=\linewidth]{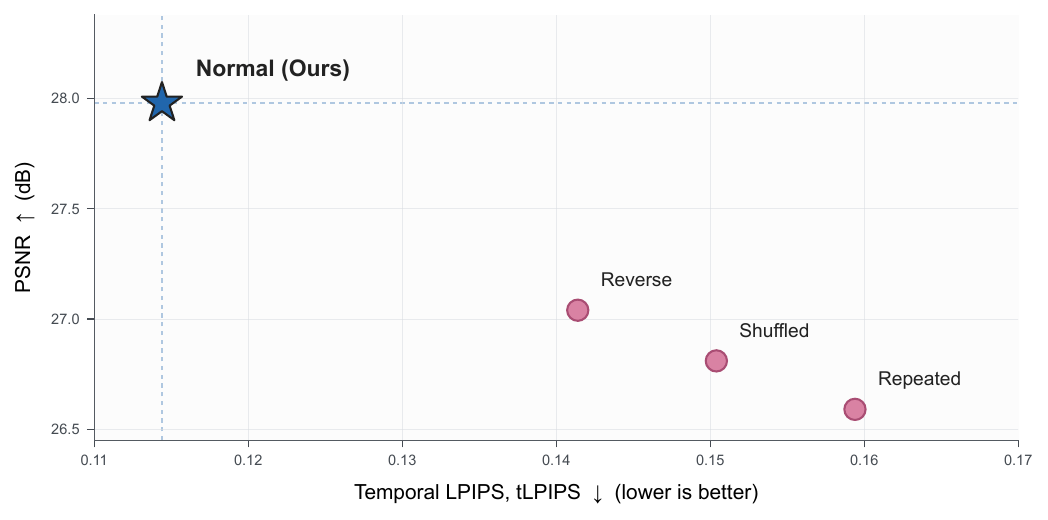}
\caption{PSNR and tLPIPS under different loop-condition orders at \(K=4\).}
\label{fig:discussion-order}
\end{figure}

Under four iterations, we compare the normal order of loop conditions with reversed, shuffled, and repeated alternatives. Figure~\visualref{fig:discussion-order} shows that the normal order attains higher PSNR and lower tLPIPS than each alternative, supporting the role of stage-specific condition ordering in iterative restoration.

\FloatBarrier
\subsection{Real-World UHD Image Restoration}\label{app:real-world-uhd}
\begin{figure}[!htbp]
\centering
\includegraphics[width=\linewidth]{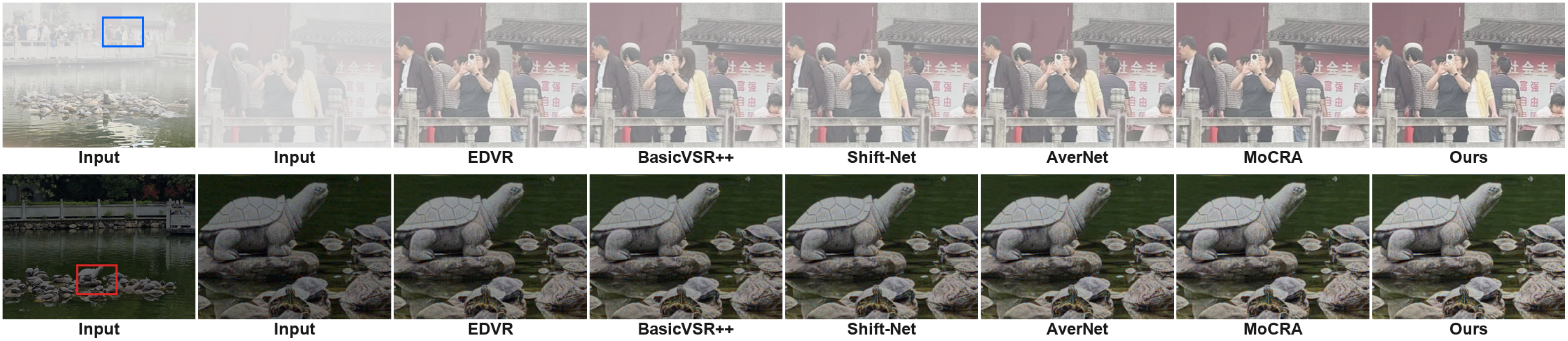}
\caption{Restoration results on real-world UHD images.}
\label{fig:real-world-uhd}
\end{figure}

Figure~\visualref{fig:real-world-uhd} presents restoration comparisons on real-world UHD images.

\end{document}